\documentclass{tlp}

\usepackage{url}
\usepackage{amsmath}
\let\proof\relax

\usepackage{amssymb,amsthm,mathrsfs,amsfonts,dsfont} 
\usepackage{xcolor,colortbl}
\usepackage{mdframed}
\usepackage{graphicx}
\usepackage{graphics} 
\usepackage{epsfig} 
\usepackage{multirow}
\usepackage{xspace}

\usepackage[utf8]{inputenc}
\usepackage{comment}
\usepackage{todonotes}
\usepackage{subcaption}
\usepackage{tabularx}
\usepackage{array}

\usepackage{todonotes}

\definecolor{asparagus}{rgb}{0.53, 0.66, 0.42}

\usepackage{tikz}
\usepackage{pgfplots}\usetikzlibrary{plotmarks}
\pgfplotsset{
    compat=1.12 % version for compiler
	, filter discard warning=false % remove notes while compiling
	, legend cell align=left
	, minor grid style={loosely dotted, lightgray}
	, major grid style={loosely dashed, lightgray}
}

\usepackage{xspace}
\usepackage{booktabs}

\usepackage{asplisting}

\theoremstyle{definition}

\theoremstyle{remark}

\begin{document}

%%%%% Per fare in modo che il contatore dei listing sia il numero di sezione
\counterwithin{lstlisting}{section}
%%%%%

\newcounter{examplecounter}
\newcommand{\examplerow}[2]
{
    \ignorespaces#2 & (\refstepcounter{examplecounter}\theexamplecounter\label{#1})\\[\smallskipamount]
}

\newcommand{\examplemultirow}[3]
{
    \stepcounter{examplecounter}\ignorespaces#2 & \multirow{#1}{*}{(\theexamplecounter)}\\\ignorespaces#3\\[\smallskipamount]
}

\newcounter{exampleblockcounter}
\counterwithin*{exampleblockcounter}{section}

\newcommand{\whichexample}{\textbf{Example~\thesection.\theexampleblockcounter.} & \\[\smallskipamount]}

\newenvironment{exampleslisting}
{
    \\[\medskipamount]
    \tabularx{\linewidth}{p{0.75\linewidth} >{\raggedleft\arraybackslash}X}
}
{
    \endtabularx
}

\newenvironment{examplestext}
{\\[\smallskipamount]}
{}

\newenvironment{examplesblock}
{\stepcounter{exampleblockcounter}}
{}
\title{CNL2ASP: converting controlled natural language sentences into ASP}

\righttitle{CNL2ASP: converting controlled natural language sentences into ASP}

\jnlPage{1}{8}
\jnlDoiYr{2021}
\doival{10.1017/xxxxx}

\begin{authgrp}
    \author{\gn{Simone} \sn{Caruso}}
    \affiliation{DIBRIS, University of Genova, Genova, Italy}
    \author{\gn{Carmine} \sn{Dodaro}}
    \affiliation{DeMaCS, University of Calabria, Rende, Italy}
    \author{\gn{Marco} \sn{Maratea}}
    \affiliation{DIBRIS, University of Genova, Genova, Italy; \& DeMaCS, University of Calabria, Rende, Italy}
    \author{\gn{Marco} \sn{Mochi}}
    \affiliation{DIBRIS, University of Genova, Genova, Italy}
    \author{\gn{Francesco} \sn{Riccio}}
    \affiliation{Engineering Division, ALTEN Italia, Torino, Italy}
\end{authgrp}
\lefttitle{Caruso et al.}

\history{\sub{xx xx xxxx;} \rev{xx xx xxxx;} \acc{xx xx xxxx}}

\maketitle

\begin{abstract}
Answer Set Programming (ASP) is a popular declarative programming language for solving hard combinatorial problems. Although ASP has gained widespread acceptance in academic and industrial contexts, there are certain user groups who may find it more advantageous to employ a higher-level language that closely resembles natural language when specifying ASP programs.
In this paper, we propose a novel tool, called CNL2ASP, for translating English sentences expressed in a controlled natural language (CNL) form into ASP.
In particular, we first provide a definition of the type of sentences allowed by our CNL and their translation as ASP rules, and then exemplify the usage of the CNL for the specification of both synthetic and real-world combinatorial problems.
Finally, we report the results of an experimental analysis conducted on the real-world problems to compare the performance of automatically generated encodings with the ones written by ASP practitioners, showing that our tool can obtain satisfactory performance on these benchmarks.
Under consideration in Theory and Practice of Logic Programming (TPLP).
\end{abstract}


{\stepcounter{exampleblockcounter}}
{}

\maketitle      % typeset the title of the contribution
\section{Introduction}
Answer Set Programming (ASP)~\citep{DBLP:books/sp/Lifschitz19,DBLP:journals/cacm/BrewkaET11,DBLP:conf/iclp/GelfondL88} is a well-known declarative programming paradigm proposed in the area of Knowledge Representation and Reasoning (KRR) and geared toward solving hard combinatorial problems.
As a matter of fact, ASP has been widely used for solving problems in both academic and industrial contexts~(see \citep{DBLP:journals/aim/ErdemGL16} for a complete survey on ASP applications). The success of ASP is mainly due to its simple syntax, its intuitive semantics, and the availability of efficient systems, like \textsc{clingo}~\citep{DBLP:conf/iclp/GebserKKOSW16} or \textsc{dlv}~\citep{DBLP:conf/lpnmr/AlvianoCDFLPRVZ17}.

Nevertheless, despite the success of ASP, and in general of KR formalisms, it may be
preferable for certain types of users to use a higher-level language that is closer to natural
language for specifying ASP programs.
For this reason, in the last decades, a number of attempts to convert English sentences expressed in a controlled natural language (CNL) into a KR formalism emerged \citep{DBLP:conf/iccs/Fuchs05,DBLP:conf/flairs/ClarkHJTW05}.
In the context of ASP, a CNL has been used for solving logic puzzles \citep{DBLP:conf/kr/BaralD12}, and for answering biomedical queries \citep{DBLP:conf/bionlp/ErdemY09}.

Arguably, using a CNL may offer several practical advantages:
\begin{enumerate}    
    \item CNL specifications are usually more readable.
    \item Writing CNL specifications is expected to be easier and faster than encoding knowledge in a formal KR language, like ASP. The generated ASP encodings can be used as a starting point for further optimization made by ASP experts.
    \item CNL specifications tend to be more adaptable to changes compared to ASP encodings, e.g., adding a term in an ASP atom requires the substitution of all occurrences of the atoms, whereas in a CNL this should have almost no impact.
    %\item CNL specifications make it easier to create explanations over complex domains~\citep{DBLP:conf/aaai/OztokE11}.
    \item CNL specifications can be used as a basis for deploying richer language processing.
\end{enumerate}

The contribution of this paper is in the aforementioned context. 
In fact, we propose a tool called CNL2ASP that automatically translates sentences expressed in a CNL language into ASP rules.
The CNL supported by CNL2ASP is inspired by the Semantics of Business Vocabulary and Business Rules (SBVR)~\citep{DBLP:conf/aaaiss/BajwaLB11,sbvr}, which is a standard proposed by the Object Management Group to formally describe complex entities, e.g., the ones related to a business, using natural language, and by PENG\textsuperscript{ASP}, a CNL defined by \cite{DBLP:journals/tplp/Schwitter18}.

The development of the tool has been oriented towards four different types of use cases, i.e., \textit{(i)} enabling the possibility of specifying ASP programs also to users that have a limited experience on ASP; \textit{(ii)} to help ASP experts to create a fast prototype of intuitive encodings which are subsequently subject to optimization; \textit{(iii)} improving the readability of ASP programs since there is a one-to-one mapping between ASP rules and English specifications; and \textit{(iv)} offering a modern tool that can be used as a basis for writing specifications in a natural language.
In particular, to show the capabilities of our CNL, we reported several synthetic and real-world use cases showing how the CNL can be indeed used for solving (complex) combinatorial problems.
Moreover, we performed an experimental analysis on the real-world use cases comparing the performance of the ASP encoding generated by CNL2ASP with the one created by ASP practitioners, showing that our tool can, in general, obtain good performances.
Subsequently, we conducted a preliminary analysis to assess the usability and readability of
the proposed CNL. 
Finally, we mention that the implementation of the tool presented in this paper is open source and publicly available at~\url{https://github.com/dodaro/cnl2asp}.

\paragraph{Contributions.}
To summarize, the main contributions of this paper are the following:
\begin{enumerate}
    \item We defined a CNL designed for solving complex combinatorial problems.
    \item We implemented CNL2ASP, a tool that automatically translates English sentences expressed in such a CNL into a corresponding ASP encoding. CNL2ASP supports the main features of the ASP language, including disjunctive and choice rules, aggregates, and weak constraints (Section~\ref{sec:cnl2asp}).
    \item We provided several use cases on well-known, synthetic domains (Section~\ref{sec:synthetic}), as well as on real-world problems described in the literature (Section~\ref{sec:realword}). Concerning the latter, we also provided the results of an experimental analysis comparing the performance of the generated encodings with the ones written by human experts.    
    \item We performed a preliminary user validation to evaluate the usability and the readability of the CNL (Section~\ref{sec:uservalidation}).
\end{enumerate}

\section{Preliminaries}
\label{sec:back}
Answer Set Programming (ASP) \citep{DBLP:journals/cacm/BrewkaET11} is a programming paradigm
developed in the area of KRR and logic programming.
In this section, we overview the language of ASP~\citep{DBLP:journals/tplp/CalimeriFGIKKLM20}.
\paragraph{Syntax.}
The syntax of ASP is similar to the one of Prolog.
Variables are strings starting with an uppercase letter, and
constants are integers or strings starting with lowercase letters.
A {\em term} is either a variable or a constant.
A {\em standard atom} is an expression \lstinline[language=asp]|p(t$_1$,$\ldots$,t$_n$)|, where \lstinline[language=asp]|p| is a
{\em predicate} of arity $n$ and \lstinline[language=asp]|t$_1$,$\ldots$,t$_n$| are terms.
An atom \lstinline[language=asp]|p(t$_1$,$\ldots$,t$_n$)| is ground if \lstinline[language=asp]|t$_1$,$\ldots$,t$_n$| are constants.
A {\em ground set} is a set of pairs of the form $\langle consts\! :\!conj \rangle$,
where $consts$ is a list of constants and $conj$ is a conjunction of ground standard atoms.
A {\em symbolic set} is a set specified syntactically as
\lstinline[language=asp]|{Terms$_1$ : Conj$_1$; $\ldots$, Terms$_t$ : Conj$_t$}|
where $t>0$, and for all $i \in [1,t]$, each \lstinline[language=asp]|Terms$_i$| is a non-empty list of terms, and each \lstinline[language=asp]|Conj$_i$| is a non-empty conjunction of standard atoms.
A {\em set term} is either a symbolic set or a ground set.
Intuitively, a set term \lstinline[language=asp]|{X:a(X,c),p(X); Y:b(Y,m)}|
stands for the union of two sets: the first one contains the \lstinline[language=asp]|X|-values making the conjunction \lstinline[language=asp]|a(X,c), p(X)| true, and the second one contains the \lstinline[language=asp]|Y|-values making the conjunction \lstinline[language=asp]|b(Y,m)| true.
An {\em aggregate function} is of the form $f(S)$, where $S$ is a
set term, and $f$ is an {\em aggregate function symbol}.
%Syntactically a set term can be expressed as union symbolic sets
%
%XXXXX
%
% of set terms,
%in that case
Basically, aggregate functions map multisets of constants to a constant.
The most common functions implemented in ASP systems are \lstinline[language=asp]|#count|, for counting number of terms; \lstinline[language=asp]|#sum|, for computing sum of integers, \lstinline[language=asp]|#min|, for computing the minimum integer in a set, and  \lstinline[language=asp]|#max|, for computing the maximum integer in a set~\citep{DBLP:journals/ai/FaberPL11}.
%
%\begin{itemize}	
%    \item \lstinline[language=asp]|#min|, minimum integer;
%	\item \lstinline[language=asp]|#max|, maximum integer;
%    \item \lstinline[language=asp]|#count|, number of terms;
%	\item \lstinline[language=asp]|#sum|, sum of integers.
%\end{itemize}
An {\em aggregate atom} is of the form $f(S) \prec T$, where $f(S)$ is an
aggregate function, \lstinline[language=asp]|$\prec\ \in \{$<, <=, >, >=, !=, =$\}$|
is a comparison operator, and $T$ is a term called guard.
An aggregate atom $f(S) \prec T$ is ground if $T$ is a constant and
$S$ is a ground set.
An \emph{atom} is either a standard atom or an aggregate atom.
A {\em rule} $r$ has the following form:

\begin{asp}[numbers=none]
	a$_1$ | $\ldots$ | a$_n$ :- b$_1$, $\ldots$, b$_k$, not b$_{k+1}$, $\ldots$, not b$_m$.
\end{asp}

\noindent where \lstinline[language=asp]|a$_1$,$\ldots$,a$_n$| are standard atoms (with $n \geq 0$), \lstinline[language=asp]|b$_1$,$\ldots$,b$_k$| are atoms, and
\lstinline[language=asp]|b$_{k+1}$,$\ldots$,b$_m$| are standard atoms (with $m\geq k \geq 0$).
A literal is either a standard atom \lstinline[language=asp]|a| or its negation \lstinline[language=asp]|not a|.
The disjunction \lstinline[language=asp]{a$_1$ | $\ldots$ | a$_n$} is the {\em head} of $r$, while
the conjunction \lstinline[language=asp]|b$_1$, $\ldots$, b$_k$, not b$_{k+1}$, $\ldots$, not b$_m$| is its {\em body}. Rules with empty body are called {\em facts}.
Rules with empty head are called {\em constraints}. % All the other ones are {\em regular} rules.
A variable that appears uniquely in set terms of a rule $r$ is said to be {\em local} in $r$, otherwise it is a {\em global} variable of $r$.
An ASP program is a set of {\em safe} rules, where
a rule $r$ is {\em safe} if the following conditions hold:
{\em (i)} for each global variable $X$ of $r$ there is a positive standard atom
$\ell$ in the body of $r$ such that $X$ appears in $\ell$; and
{\em (ii)} each local variable of $r$ appearing in a symbolic set
\lstinline[language=asp]|{Terms : Conj}| also appears in a positive atom in \lstinline[language=asp]|Conj|.
A {\em weak constraint} $\omega$~\citep{DBLP:journals/tkde/BuccafurriLR00} is of the form:

\begin{asp}[numbers=none]
    :~ b$_1$, $\ldots$, b$_k$, not b$_{k+1}$, $\ldots$, not b$_m$. [w@l, t$_1$, $\ldots$, t$_z$]
\end{asp}

\noindent where \lstinline[language=asp]|t$_1$, $\ldots$, t$_z$| are terms, \lstinline[language=asp]|w| and \lstinline[language=asp]|l| are the weight and level of $\omega$, respectively.
Intuitively, \lstinline[language=asp]|[w@l]| is read ``as weight \lstinline[language=asp]|w| at level \lstinline[language=asp]|l|'', 
where weight is the ``cost'' of violating the condition in the body of \lstinline[language=asp]|w|, 
whereas levels can be specified for defining a priority among preference criteria.
An ASP program with weak constraints is $\Pi = \langle P,W \rangle$, where $P$ is a program
and $W$ is a set of weak constraints.
A standard atom, a literal, a rule, a program or a weak constraint is {\em ground} if no variables appear in it.

\paragraph{Semantics.}
Let $P$ be an ASP program. The {\em Herbrand universe} $U_{P}$ and
the {\em Herbrand base} $B _{P}$ of $P$ are defined as usual.
%
%of $P$, denoted as $U_{P}$, is the set of all constants appearing in $P$.
%The {\em Herbrand base} of $P$,
%denoted as $B _{P}$, is the set of all ground atoms constructible
%from the predicate symbols appearing in $P$ and the constants of $U_{P}$.
The ground instantiation $G_P$ of $P$ is the set of all the ground instances of rules of $P$ that can be obtained by substituting variables with constants from $U_{P}$.
An {\em interpretation} $I$ for $P$ is a subset $I$ of $B_{P}$.
A ground standard atom \lstinline[language=asp]|p| is true w.r.t. $I$
if \lstinline[language=asp]|p $\in I$|, and false otherwise.
A literal \lstinline[language=asp]|not p| is true w.r.t. $I$ if \lstinline[language=asp]|p| is false w.r.t. $I$, and false otherwise.
An aggregate atom is true w.r.t. $I$ if the evaluation of its aggregate function
(i.e., the result of the application of $f$ on the multiset $S$) with respect to $I$
satisfies the guard; otherwise, it is false.
A ground rule $r$ is {\em satisfied} by $I$
if at least one atom in the head is true w.r.t. $I$ whenever all conjuncts of the body
of $r$ are true w.r.t. $I$.
A model is an interpretation that satisfies all rules of a program.
Given a ground program  $G_P$ and an interpretation  $I$, the
{\em reduct} of $G_P$ w.r.t. $I$ is the subset $G_P^I$ of $G_P$ obtained
by deleting from $G_P$ the rules in which a body literal is false w.r.t. $I$~\citep{DBLP:journals/ai/FaberPL11}.
An interpretation $I$ for $P$ is an {\em answer set} (or stable model)
for  $P$ if $I$ is a minimal model (under subset inclusion) of $G_P^I$
(i.e.,  $I$ is a minimal model for $G_P^I$).
Given a program with weak constraints $\Pi = \langle P,W \rangle$ and an interpretation $I$, the semantics of $\Pi$ extends from the basic case defined above. Thus, let $G_P$ and $G_W$ be the instantiation of $P$ and $W$, respectively. Then, let $G^I_W$ be the set
\begin{asp}[numbers=none]
$\{($w@l,t$_1$,$\ldots$,t$_z)\ \mid\ $:~ b$_1$,$\ldots$,b$_k$,not b$_{k+1}$,$\ldots$,not b$_m$. [w@l,t$_1$,$\ldots$,t$_z$] $\in G_W$, $b_1$,$\ldots$,b$_k$ $\in I $, $and \ $b$_{k+1}$,$\ldots$,b$_m$ $\not\in I\}.$
\end{asp}
Moreover, for an integer $l$, $P_l^I = \sum_{(w@l, t_1,\ldots,t_z) \in G^I_W} w$ if there is at least one tuple in $G^I_W$ whose level is equal to $l$, and 0 otherwise.
Given a program with weak constraints $\Pi = \langle P,W \rangle$, an answer set $M$ for $P$ is said to be dominated by an answer set $M'$ for $P$, if there exists an integer $l$ such that $P_l^{M'} < P_l^M$ and $P_{l'}^{M'} = P_{l'}^M$ for all integers $l' > l$.
An answer set $M$ for $P$ is said to be \textit{optimal} or \textit{optimum} for $\Pi$ if there is no other answer set $M'$ that dominates $M$~\citep{DBLP:journals/tplp/CalimeriFGIKKLM20}.

\paragraph{Syntactic shortcuts.}
In the following, \lstinline[language=asp]|p(1..n).| denotes the set of facts \lstinline[language=asp]|p(1).| \lstinline[language=asp]|$\ldots$ p(n).|
Moreover, we use \textit{choice rules} of the form \lstinline[language=asp]|{X}|, where \lstinline[language=asp]|X| is a set of atoms. Choice rules of this kind can be viewed as a syntactic shortcut for the rule 
\lstinline[language=asp]{p | p'.} for each \lstinline[language=asp]|p| $\in$ \lstinline[language=asp]|X|, where \lstinline[language=asp]|p'| is a fresh new atom not appearing elsewhere in the program, meaning that the atom $p$ can be chosen as true.
Choice rules can also have bounds, i.e., \lstinline[language=asp]|1 <= {X} <= 1|, and in this case can be seen as a shortcut for the choice rule \lstinline[language=asp]|{X}| and the rule \lstinline[language=asp]|:- #count{X} != 1.|

\section{CNL2ASP}\label{sec:cnl2asp}
\begin{figure}[t!]
    \centering
    \includegraphics[width=\textwidth]{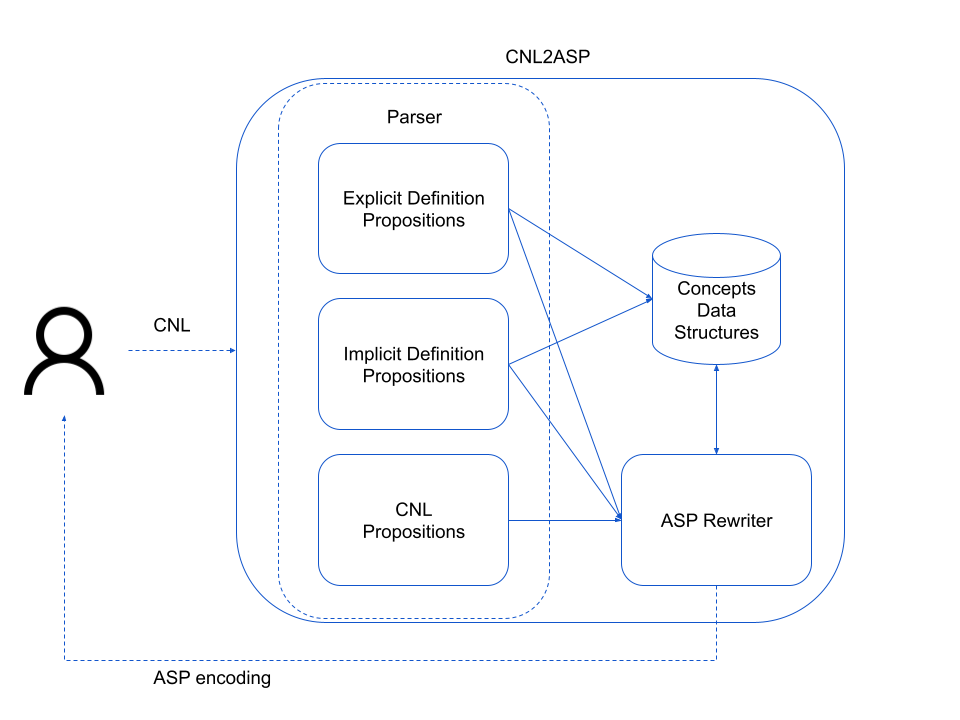}
    \caption{Architecture of the tool CNL2ASP.}
    \label{fig:architecture}
\end{figure}
This section deals with the specification of CNL language and with the implementation of the tool CNL2ASP, whose architecture is depicted in Figure~\ref{fig:architecture}.
The tool takes as input a file containing a list of statements written in a CNL and produces as output a file containing a set of ASP rules.
A specification written in this CNL is made of propositions, the structure of which is defined by clauses, linked by connectives, that are used to express concepts, to query them for information or to express conditions on them.
Concepts in a proposition define the application domain, i.e., they describe entities that are used as subjects of other propositions.
The combination of clauses that produces a proposition defines its type, that is used to understand what the proposition is supposed to mean and how that meaning can be translated into ASP rules and facts.

CNL2ASP is made of three main components, namely the \textit{Parser}, the \textit{Concepts Data Structures}, and the \textit{ASP Rewriter}.
In particular, each CNL proposition in the input file is processed by the Parser, whose role is \textit{(i)} to create appropriate data structures for concepts to be stored in the Concept Data Structures, and \textit{(ii)} to tokenize the CNL statements and send the result to the ASP Rewriter component.
In more details, the Parser interprets three subtypes of CNL propositions, namely explicit definition propositions, implicit definition propositions, and (standard) CNL propositions.
In particular, the starting production rule is the following:
\begin{cnl}
start $\longrightarrow$ (explicit_definition_proposition | implicit_definition_proposition | standard_proposition)+
\end{cnl}
The first two types of propositions are used to define the concepts, where in our context a concept is a thing, a place, a person or an object that is used to model entities of the application domain of the CNL.
Standard CNL propositions are sentences describing the rules of the application domain.
As an example, consider the application domain of describing the rights and the obligations of a customer of an online store.
In this context, concepts can be the customer, the company, the product, and so on, whereas CNL propositions are sentences stating what actions customers and companies can/cannot do.
It is important to highlight that, in our tool, concepts are exclusively defined by their names. Consequently, taking the earlier example into account, there exists only a single concept for customer, company, product, and so forth.
After all the sentences have been processed by the Parser, they are sorted as follows: (explicit and implicit) definition propositions are processed before (standard) CNL propositions.
Among the CNL propositions, the ASP Rewriter first processes the ones that are related to choice and disjunctive rules since they can also define new concepts in the data structures, and then processes strong and weak constraints.
For each proposition, the ASP Rewriter first initializes the ASP atoms, then creates aggregates, arithmetic operations and comparisons, and further it merges all of them to create the head and the body of the ASP rules.
Finally, after all ASP rules are created, they are stored in a output file that is returned to the user.
In the following sections we first describe the different propositions accepted by the Parser along with their grammar\footnote{For the sake of readability, we only provide basic elements of the grammar, we refer the reader to \citep{grammar} for the full grammar.} and their translation as ASP rules (Sections~\ref{sec:explicit},~\ref{sec:implicit}, and~\ref{sec:standard}), and then we report a brief description of the usage of the tool (Section \ref{sec:usage}).

\subsection{Explicit definition propositions}\label{sec:explicit}
Explicit definition propositions are used to define the concepts occurring in the domain application, and they are used to create data structures which are later on used by the ASP Rewriter to produce ASP rules.
In more details, the production rule of explicit definition propositions is the following:
\begin{cnl}
explicit_definition_proposition $\longrightarrow$ (domain_definition | temporal_concept_definition)CNL_END_OF_LINE
\end{cnl}
where each proposition is terminated by \lstinline[language=cnl]|CNL_END_OF_LINE| (in our case, a dot).
In particular, an explicit definition proposition can be either a \lstinline[language=cnl]|domain_definition|, used to define all the entities of the problem and their structure; or a \lstinline[language=cnl]|temporal_concept_definition|, used to define only temporal elements, as days or timeslots.

\subsubsection{Domain definition}
Domain definitions start with a subject optionally followed by the sentence \lstinline[language=cnl]|"is identified by"| and the definition of the keys, i.e., the parameters that uniquely represent the entity and then, also optionally, a sentence used to express the other parameters. 
The production rule is the following:
\begin{cnl}
domain_definition: ("A " | "An ")? subject_name ("is identified by" atom_key)? ", and"? ("has" parameter ((","|", and") parameter)*)?
\end{cnl}
Domain definitions are not directly translated as ASP rules, instead they are used to add elements in the data structures.
All properties can be later on used in propositions to refer specific properties of a concept. By default, if no property is referred to by a sentence, then the identifier is used.

The following sentences are examples of domain definitions:
\begin{cnl}
A movie is identified by an id, and has a title, a director, and a year.
A director is identified by a name.
A topMovie is identified by an id.
A scoreAssignment is identified by a movie, and by a value.
\end{cnl}
Note that \lstinline[language=cnl]|scoreAssignment| is identified by a movie, which is a concept that is created by the user. This has an impact on its translation into ASP, as shown in Section~\ref{sec:wheneverthen}.

\subsubsection{Temporal concept definition}\label{sec:temporalconcept}
Temporal concept definitions start with a subject followed by the sentence \lstinline[language=cnl]|"is a temporal concept expressed in"|, then by the temporal type that can be minutes, days or steps. The preposition continues with a sentence used to express the temporal range and, finally, it can be closed with a sentence used to specify the length of each temporal step.
The production rule is the following:
\begin{cnl}
temporal_concept_definition: ("A " | "An ") subject_name "is a temporal concept expressed in" CNL_TEMPORAL_TYPE "ranging from" temporal_range_start "to" temporal_range_end ("with a length of" CNL_NUMBER ("minutes" | "days"))?
\end{cnl}
Temporal concepts enable the possibility to refer to them using special words like after, before, and so on.

An example of a temporal definition is the following sentence:
\begin{cnl}
A timeslot is a temporal concept expressed in minutes ranging from 07:00 AM to 09:00 AM with a length of 30 minutes.
\end{cnl}
Such concepts are conveniently translated as ASP facts by the ASP Rewriter as follows:
\begin{asp}
timeslot(1,"07:00").
timeslot(2,"07:30").
timeslot(3,"08:00").
timeslot(4,"08:30").  
\end{asp}
and the association between the used number and the corresponding time slot is stored into a dedicated data structure, so that when a user refers to a particular time slot (e.g., 07:30 AM), it is automatically encoded as the corresponding ASP atom (e.g., \lstinline[language=asp]|timeslot(2, "07:30")|).
The second term is a string representing the time slot, which is never used in the generated ASP encoding, but that can be useful when provided as output to the user.

\subsection{Implicit definition propositions}\label{sec:implicit}
Implicit definition propositions are used to define concepts that can, then, be used by other propositions. These definitions express the \emph{signature} of the concept indicated by the subject of the proposition, carrying information regarding the concept in the definition that our tool can use later on in the specification whenever the same concept is used.
Differently from explicit definition propositions, users do not have to specify the properties of the concepts, because they are inferred from the sentence.
In more details, the production rule of implicit definition propositions is the following:
\begin{cnl}
implicit_definition_proposition $\longrightarrow$ (constant_definition_clause | compounded_clause | enumerative_definition_clause)CNL_END_OF_LINE
\end{cnl}
In particular, an implicit definition proposition can be a \lstinline[language=cnl]|constant_definition_clause|, used to specify constants;
or a \lstinline[language=cnl]|compounded_clause|, used to define elements using lists and ranges;
or a \lstinline[language=cnl]|enumerative_definition_clause|, used to define elements one at a time, optionally closing the proposition with a \lstinline[language=cnl]|when| clause, defining a condition in which the element is defined (e.g., \lstinline[language=cnl]|X is true when Y is true|), and with a \lstinline[language=cnl]|where| clause.

\subsubsection{Constant definitions}\label{sec:constantdefinitions}
Constant definitions are used to introduce constants to be used later on in the specification.

The following sentences are examples of constant definitions:
\begin{cnl}
minKelvinTemperature is a constant equal to 0.$\label{prop:constantequalto}$
acceptableTemperature is a constant.$\label{prop:constant}$
\end{cnl}
As we can see from the proposition at line \ref{prop:constantequalto}, the constant 0 is introduced with a \lstinline[language=cnl]|equal to| clause, and it is bound to the subject of the proposition.
Instead, in the proposition at line \ref{prop:constant}, we are defining a constant without assigning it a value, which can be later on assigned by the user (e.g., the ASP system \textsc{clingo}~\citep{DBLP:conf/iclp/GebserKKOSW16} supports the option \lstinline[language=cnl]|--const| to specify constants).
In the case of constant definitions, there are no translations to ASP available, because they are instead stored in the data structures and substituted in the resulting program when the subject of the definition is used.

\subsubsection{Compound definitions}
Compound definitions are used to introduce a set of related concepts all at once, by making use of either ranges of numbers or lists.
The following sentences are examples of compound definitions:
\begin{cnl}
A ColdTemperature goes from minKelvinTemperature to acceptableTemperature.$\label{prop:coldtemperature}$
A day goes from 1 to 365.$\label{prop:days}$
A drink is one of alcoholic, nonalcoholic and has color that is equal to respectively blue, yellow.$\label{prop:drink}$
\end{cnl}

Propositions at lines \ref{prop:coldtemperature} and \ref{prop:days} are examples of definitions using a range, identified by the construct \lstinline[language=cnl]|goes from/to|. In particular, proposition at line \ref{prop:coldtemperature} uses the constants defined in Section~\ref{sec:constantdefinitions}.

Proposition at line \ref{prop:drink} is an example of a definition of a drink using lists with a \lstinline[language=cnl]|one of| clause, where one can also specify additional attributes for each element of the list in a positional way using a \lstinline[language=cnl]|respectively| clause, and a list with the same number of elements of the list enumerating all the possible values that the subject of the proposition can have.

The corresponding ASP code, in this case, is quite straightforward and is depicted below:
\begin{asp}
coldtemperature(0..acceptableTemperature).
day(1..365).
drink(1, "alcoholic", "blue").
drink(2, "nonalcoholic", "yellow").
\end{asp}
First of all, note that constant \lstinline[language=cnl]|minKelvinTemperature| is directly replaced by its value (i.e., 0), whereas constant \lstinline[language=cnl]|acceptableTemperature| is left as is.
List elements defined in proposition at line \ref{prop:drink} carry on their position number with them, which turns out to be handy as a basic way to encode precedence relationships when the subject is not a number.

\subsubsection{Enumerative definitions}
Enumerative definitions are used to introduce a property for a single concept or a relationship among a set of concepts. The peculiarity of this kind of propositions lies in the different translations into ASP as the clauses used within them change.

The following sentences are examples of enumerative definitions:
\begin{cnl}
John is a waiter.$\label{prop:john}$
1 is a pub.$\label{prop:1pub}$
Alice is a patron.$\label{prop:alice}$
Waiter John works in pub 1.$\label{prop:johnworks}$
Waiter John serves a drink alcoholic.$\label{prop:serves}$
Pub 1 is close to pub 2 and pub X, where X is one of 3,4.$\label{prop:pub1close}$
Waiter W is working when waiter W serves a drink.$\label{prop:workswhenserves}$
\end{cnl}
Such propositions show the construction to define relationships or properties related to a particular subject.
In particular, propositions from line $\ref{prop:john}$ to line $\ref{prop:alice}$ are used to define the concepts of waiter, pub, and patron, respectively, whereas propositions at lines \ref{prop:johnworks} and \ref{prop:serves} define concepts related to work in and to serve, respectively.
Proposition at line \ref{prop:pub1close} illustrates another feature of our CNL, i.e., \lstinline[language=cnl]|where| clauses, that are used in the example to define the values that the variable \lstinline[language=cnl]|X| can take.
Proposition at line \ref{prop:workswhenserves} is a \textit{conditional} definition, identified by a \lstinline[language=cnl]|when| clause.

The translations in ASP of these examples are the following:
\begin{asp}
waiter("john").
pub(1).
patron("alice").
work_in("john", 1).
serve("john", "alcoholic").
close_to(1, 2, 3).  close_to(1, 2, 4).
working(W) :- serve(W,_).
\end{asp}
Propositions from line \ref{prop:john} to line \ref{prop:pub1close} always hold true, therefore they are used by the ASP Rewriter to produce the corresponding ASP facts. In particular, proposition at line \ref{prop:pub1close} is translated in a similar manner to compound definitions with lists.
Instead, proposition at line \ref{prop:workswhenserves} holds true only if the statement introduced by the \lstinline[language=cnl]|when| clause is true, hence it is translated into an ASP rule, where the body of the rule is the element inside the \lstinline[language=cnl]|when| clause.

Note that, in these examples, \lstinline[language=cnl]|W| is considered as a variable, whereas \lstinline[language=cnl]|John| and \lstinline[language=cnl]|Alice| are treated as ASP strings. This is because CNL2ASP assumes that every object starting with an upper case letter and containing only upper case letters, numbers or symbols is considered as a variable, while other objects are strings or numbers (e.g., \lstinline[language=cnl]|MY_VARIABLE| is considered as a variable, whereas \lstinline[language=cnl]|My_String| is considered as a string).

\subsection{CNL propositions}\label{sec:standard}
Explicit and implicit definition propositions are used to define the concepts of the domain application, whose specifications are instead described by (standard) CNL propositions.
The production rule of standard CNL propositions is the following:
\begin{cnl} 
standard_proposition $\longrightarrow$ (
        whenever_then_clause_proposition |
        fact_proposition | 
        quantified_choice_proposition |
        negative_strong_constraint_proposition | 
        positive_strong_constraint_proposition | 
        weak_constraint_proposition |         
        )CNL_END_OF_LINE
\end{cnl}
Therefore, the CNL considers several types of propositions, which are described in the following.

\subsubsection{Whenever/then clauses}\label{sec:wheneverthen}
\lstinline[language=cnl]|Whenever/then| clauses are used to describe actions occurring when a condition is fulfilled. In more details, the production rule is the following:
\begin{cnl}
whenever_then_clause $\longrightarrow$ (whenever_clause ","?)+ then_clause     
\end{cnl}
They start with \lstinline[language=cnl]|whenever| clauses, i.e., sentences specifying conditions, followed by a \lstinline[language=cnl]|then| clause, that is a sentence used to express the actions that must or can hold when the \lstinline[language=cnl]|whenever| clauses are fulfilled.

The following sentences are examples of \lstinline[language=cnl]|whenever/then| clauses:
\begin{cnl}
Whenever there is a movie with director equal to spielberg, with id X then we must have a topmovie with id X.
Whenever there is a director with name X different from spielberg then we can have at most 1 topmovie with id I such that there is a movie with director X, and with id I.
Whenever there is a movie with id I, with director equal to nolan then we can have a scoreAssignment with movie I, and with value equal to 3 or a scoreAssignment with movie I, and with value equal to 2.
\end{cnl}
Such propositions are encoded in ASP as follows:
\begin{asp}
topmovie(X) :- movie(X,_,"spielberg",_).
0 <= {topmovie(I):movie(I,_,X,_)} <= 1 :- director(X), X != "spielberg".
scoreassignment(movie(I),3) | scoreassignment(movie(I),2) :- movie(I,_,"nolan",_).
\end{asp}
In particular, the form \lstinline[language=cnl]|whenever/then| followed by the word \lstinline[language=cnl]|must| is translated as a normal rule by the ASP Rewriter, whereas if it is followed by the word \lstinline[language=cnl]|can| then it can be translated as a choice rule or as a disjunctive rule depending on whether the CNL sentence contains the keyword \lstinline[language=asp]|or|.
Here, we want also to emphasize the fact that the first term of \lstinline[language=cnl]|scoreAssignment| is of the form \lstinline[language=asp]|movie(I)| since it is defined to be of the type movie.

\subsubsection{Fact proposition}
Fact propositions are used to define the facts of the problem. Differently from implicit definition propositions, here no new concepts are introduced, meaning that all concepts used here must be explicitly defined.
An example of a fact proposition is the following sentence:
\begin{cnl}
There is a movie with id equal to 1, with director equal to spielberg, with title equal to jurassicPark, with year equal to 1993.
\end{cnl}
This sentence is translated as:
\begin{asp}
movie(1,"jurassicPark","spielberg",1993).
\end{asp}
It is worth mentioning that the order of the elements listed in the sentence has no impact on its translation, since the properties of the concepts are explicitly defined. Therefore, the specifications listed below all produce the same ASP output.
\begin{cnl}
There is a movie with id equal to 1, with director equal to spielberg, with year equal to 1993, with title equal to jurassicPark.
There is a movie with director equal to spielberg, with year equal to 1993, with id equal to 1, with title equal to jurassicPark.
\end{cnl}

\subsubsection{Quantified choice propositions}
Quantified choice propositions are used to define relationships or properties that \emph{can} be true for a given set of selected concepts following a choice. Also these propositions define a \emph{signature} for the concept upon which the choice has to be made.
Quantified propositions are always introduced by the \lstinline[language=cnl]|every| quantifier and, since they express possibilities, always contain a \lstinline[language=cnl]|can| clause.
In more details, the production rule is the following:
\begin{cnl}
quantified_choice_proposition $\longrightarrow$ quantifier subject_clause "can" CNL_COPULA? (verb_name | verb_name_with_preposition) (quantified_exact_quantity_clause | quantified_range_clause)? (quantified_object_clause | disjunctive_clause)? foreach_clause?
\end{cnl}
Thus, they start with a quantifier, and are always followed by a subject and a verb, optionally connected by a \lstinline[language=cnl]|CNL_COPULA| (e.g., \lstinline[language=cnl]|is, is a, is an, ...|) and then, also optionally, either by a sentence of type \lstinline[language=cnl]|quantified_exact_quantity_clause|, used to express the quantity in exact terms (e.g., \lstinline[language=cnl]|exactly 1|); or by a sentence of type  \lstinline[language=cnl]|quantified_range_clause|, used to express it using a range (e.g., \lstinline[language=cnl]|between 1 and 2|).
The proposition can be closed either with an object clause, i.e., a sentence used to express an object for the proposition, in a \lstinline[language=cnl]|subject verb object| fashion, or with a disjunctive clause; and, finally, a \lstinline[language=cnl]|foreach| clause, i.e., a sentence used to express additional objects for which any possible value can be tried.

The following sentences are examples of quantified choice propositions:
\begin{cnl}
Every patron can drink in exactly 1 pub for each day. $\label{prop:candrinkexactly1}$
Every waiter can serve a drink. $\label{prop:canserveadrink}$
Every movie with id I can have a scoreAssignment with movie I, and with value equal to 1 or a scoreAssignment with movie I, and with value equal to 2, or a scoreAssignment with movie I, and with value equal to 3. $\label{prop:disjunction}$
\end{cnl}
Proposition at line \ref{prop:candrinkexactly1} shows how one can express an exact number of choices that can be made for the concept expressed by the subject, and also how other concepts can be used in tandem with the subject to create a sort of cartesian product of choices, using a \lstinline[language=cnl]|for each| clause. These last constructions are optional, as shown in proposition at line \ref{prop:canserveadrink}.
Proposition at line \ref{prop:disjunction}, instead, shows an example of a disjunctive clause.
Their full translations into ASP is shown below:
\begin{asp}[numbers=none]
1 <= {drink_in(_X1,_X2,_X3):pub(_X3)} <= 1 :- patron(_X1), day(_X2).
{serve(_X1,_X2):drink(_,_X2,_)} :- waiter(_X1).
scoreassignment(movie(I),1) | scoreassignment(movie(I),2) | scoreassignment(movie(I),3) :- movie(I,_,_,_).
\end{asp}
The first two translations use choice rules (possibly with bounds), that are the ASP constructs that make it possible to represent propositions of this type, whereas the third one uses a disjunctive rule.
Note that the first two rules also employ generated variables (starting with symbol \lstinline[language=asp]|_|) that are used wherever two atoms have to be bound and no variable to use has been found in the specification given in input. This feature enables the specification writer to avoid cluttering the document with unnecessary variables, as can be seen throughout the propositions, with the only limitation that anaphoras have to be expressed explicitly by providing the correct variable.

\subsubsection{Negative and positive strong constraints}
Negative and positive strong constraint propositions are used to define assertions that \emph{must} be true for a given set of selected concepts. This kind of propositions does not introduce new \emph{signatures} but, on the contrary, they consume other signatures that were previously defined, meaning that the concepts used inside such constraints have to be defined before they are used. A strong constraint can represent either a prohibition (sentences starting with \lstinline[language=cnl]|It is prohibited|) or a requirement (sentences starting with \lstinline[language=cnl]|It is required|).
After specifying if the strong constraint is a prohibition or a requirement, then a user can add \textit{simple} clauses, that are made of a subject, a verb, and related object clauses; \textit{aggregate} clauses, either in active or passive form, that define an aggregation of the set of concepts that satisfy the statement in their body with the operator that was specified (\lstinline[language=cnl]|number|, \lstinline[language=cnl]|total|, \lstinline[language=cnl]|lowest|, \lstinline[language=cnl]|highest|); or other complex clauses as shown below.

In more details, the production rules of strong constraints are the following:
\begin{cnl}
negative_strong_constraint_clause $\longrightarrow$ "it is prohibited that" (simple_clause ("and also" simple_clause)* (where_clause)? ("," (whenever_clause ","?)+)? | aggregate_clause comparison_clause (where_clause)? ("," (whenever_clause ","?)+)? | when_then_clause (where_clause)? | quantified_constraint (where_clause)? | condition_clause  "," (whenever_clause ","?)+ |  temporal_constraint "," (whenever_clause ","?)+)
positive_strong_constraint_proposition $\longrightarrow$ "it is required that" (simple_clause "," (whenever_clause ","?)+ | aggregate_clause comparison_clause (where_clause)? ("," (whenever_clause ","?)+)? | when_then_clause (where_clause)? | quantified_constraint (where_clause)? | condition_clause  "," (whenever_clause ","?)+ | temporal_constraint "," (whenever_clause ","?)+)
\end{cnl}
It is possible to observe that they start with the sentence \lstinline[language=cnl]|it is prohibited that| or with the sentence \lstinline[language=cnl]|it is required that| and are always followed 
by a simple clause, i.e., a sentence of the form \lstinline[language=cnl]|subject verb object|;
by an aggregate clause, a sentence expressing a form of aggregations (e.g., \lstinline[language=cnl]|the number of|);
by a whenever clause, described in Section~\ref{sec:wheneverthen};
by a quantified constraint, used to specify clauses with quantifiers as \lstinline[language=cnl]|every| or \lstinline[language=cnl]|any|;
or by a temporal constraint, used to specify constraints on temporal concepts as \lstinline[language=cnl]|after 11:00 AM| or \lstinline[language=cnl]|before 11:00 AM|.
After simple clauses, aggregate clauses, and quantified constraints, additional sentences can be added, which can be of the type \lstinline[language=cnl]|where_clause|, used to specify conditions; or of the type \lstinline[language=cnl]|comparison_clause|, used to specify comparison between elements (e.g., \lstinline[language=cnl]|X is equal to 1|).
%Note that negative strong constraints propositions can also have a combination of simple clauses (sentences introduced by the words \lstinline[language=cnl]|and also|), whereas positive strong constraint propositions are limited to only one simple clause, since they must be converted into ASP constraints by negating the condition expressed by the simple clause.

The following sentences are examples of negative and positive strong constraints:
\begin{cnl}
It is prohibited that waiter W1 work in pub P1 and also waiter W2 work in pub P1, where W1 is different from W2.$\label{prop:prohibitedandalso}$
It is prohibited that X is equal to Y, whenever there is a movie with id X, and with year equal to 1964, whenever there is a topMovie with id Y.$\label{prop:prohibitedwhenever}$
It is prohibited that the lowest value of a scoreAssignment with movie X is equal to 1, whenever there is a topMovie with id X. $\label{prop:prohibitedmin}$
It is required that the total value of a scoreAssignment with movie X is equal to 10, such that there is a topMovie with id X. $\label{prop:requiredsum}$
It is required that the number of pub where a waiter work in is less than 2.$\label{prop:requiredcount}$
It is required that when waiter X works in pub P1 then waiter X does not work in pub P2, where P1 is different from P2.$\label{prop:requiredwhen}$
It is required that V is equal to 3, whenever there is a movie with id I, and with director equal to spielberg, whenever there is a scoreAssignment with movie I, and with value V.$\label{prop:whenever}$
It is required that every waiter is payed. $\label{prop:requiredpayed}$
\end{cnl}
Proposition at line \ref{prop:prohibitedandalso} shows a practical example of the combination of several simple clauses, and the feature enabled by \lstinline[language=cnl]|where| clauses, that makes it possible to express comparisons between variables.
Proposition at line \ref{prop:prohibitedwhenever} shows an example of \lstinline[language=cnl]|whenever| clause.
Propositions at line \ref{prop:prohibitedmin}, at line \ref{prop:requiredsum}, and at line \ref{prop:requiredcount} show examples of aggregation expressing conditions on the minimum value, on the sum of values, and on the number of occurrences, respectively.
Proposition at line \ref{prop:requiredwhen} shows a \lstinline[language=cnl]|when/then| clause.
Proposition at line \ref{prop:whenever} shows an example of \lstinline[language=cnl]|whenever| clause in the context of positive strong constraints.
Lastly, proposition at line \ref{prop:requiredpayed} is an example of how to specify a requirement that must hold for all the elements present in a particular set of concepts.
Such propositions are encoded as ASP rules as follows:
\begin{asp}
:- work_in(W1,P1), work_in(W2,P1), W1 != W2.
:- movie(X,_,_,1964), topmovie(Y), X = Y.
:- topmovie(X), #min{_X1: scoreassignment(movie(X),_X1)} = 1.
:- #sum{_X1: scoreassignment(movie(X),_X1), topmovie(X)} != 10.
:- waiter(_X1), #count{_X2: work_in(_X1,_X2)} >= 3.
:- work_in(X,P1), work_in(X,P2), P1 != P2.
:- movie(I,_,"spielberg",_), scoreassignment(movie(I),V), V != 3.
:- not payed(_X1), waiter(_X1).
\end{asp}
Note that their translation is quite intuitive, and positive strong constraints are translated as ASP constraints by negating the condition expressed by the sentence.

\subsubsection{Weak constraint propositions}
Weak constraint propositions are used to define assertions that are \emph{preferably} true for a given set of selected concepts. Also this type of propositions consumes \emph{signatures} from previously defined concepts. They are always introduced by \lstinline[language=cnl]|it is preferred| and need the specification of the optimization objective (either minimization or maximization), and the level of priority of the optimization (low, medium or high).
The production rule is the following:
\begin{cnl}
weak_constraint_proposition $\longrightarrow$ "it is preferred that" CNL_WEAK_OPTIMIZATION_OPERATOR? ","? weak_priority_operator ","? "that" (condition_operation | aggregate_clause | subject_clause CNL_COPULA (verb_name | verb_name_with_preposition) object_clause "," whenever_clause) weak_optimization_operator? (where_clause)?
\end{cnl}
In particular, they start with the sentence \lstinline[language=cnl]|it is preferred ... that|, and can be followed by a sentence expressing the nature of the optimization (i.e., \lstinline[language=cnl]|as much as possible| or \lstinline[language=cnl]|as little as possible|), and are always followed by a priority operator, i.e., a sentence expressing the level of relevance of the constraint with respect to other weak constraints (e.g., \lstinline[language=cnl]|"with low priority"|) and either a clause followed by a whenever clause, an aggregate clause or a condition operation, i.e., a sentence expressing operations between variables in the proposition (e.g., \lstinline[language=cnl]|the sum of X and Y|). The proposition is closed with an optimization operator, i.e., a sentence expressing the nature of the optimization (i.e., \lstinline[language=cnl]|"is minimized"| or \lstinline[language=cnl]|"is maximized"|) and an optional where clause.
Note that here we have two ways for expressing the object, either in the form of as much (little) as possible at the beginning of the sentence or using is maximized (minimized) at the end of the sentence.
The two ways are equivalent, but we support both of them to make sentences more natural.
Moreover, sentences containing both kind of specifications are well-formed, thus they are correctly parsed even if they are in contrast (e.g., a user can specify \lstinline[language=cnl]|as much as possible| and \lstinline[language=cnl]|"is minimized"| in the same sentence).
However, CNL2ASP subsequently checks if this happens and, in case, it triggers an error so that only one of the form is used.

The following sentences are examples of weak constraint propositions:
\begin{cnl}
It is preferred with low priority that the number of drinks that are serve is maximized.$\label{prop:servemaximized}$
It is preferred as little as possible, with high priority, that V is equal to 1, whenever there is a scoreAssignment with movie I, and with value V, whenever there is a topMovie with id I.$\label{prop:scoreminimized}$
It is preferred, with medium priority, that whenever there is a topMovie with id I, whenever there is a scoreAssignment with movie I, and with value V, V is maximized.$\label{prop:topmoviemaximized}$
It is preferred, with medium priority, that the total value of a scoreAssignment is maximized.$\label{prop:summinimized}$
\end{cnl}
The sentence at line~\ref{prop:servemaximized} shows an example of a maximization over the result of a \lstinline[language=asp]|#count| aggregate.
The sentence at line~\ref{prop:scoreminimized} instead is an example of minimization using the form \lstinline[language=cnl]|as little as possible|.
Then, the sentence at line~\ref{prop:topmoviemaximized} shows a sentence where the subject of maximization is a variable defined in \lstinline[language=cnl]|scoreAssignment|.
Finally, the sentence at line~\ref{prop:summinimized} is an example of a \lstinline[language=asp]|#sum| aggregate, where the result of the aggregation is subject to maximization.
Translation of the propositions above are shown below:
\begin{asp}
:~ #count{_X1: serve(_,_X1)} = _X2. [-_X2@1]
:~ scoreassignment(movie(I),V), topmovie(I), V = 1. [1@3, I,V]
:~ topmovie(I), scoreassignment(movie(I),V). [-V@2, I]
:~ #sum{_X1: scoreassignment(movie(_),_X1)} = _X2. [-_X2@2]
\end{asp}
Also in this case the translation is quite intuitive, however one should note that maximization constructs are translated using negative weights.

\subsection{Usage}\label{sec:usage}
In this section, we provide a few technical details and report the usage of the tool.
CNL2ASP has been implemented using the programming language Python, and the open-source library lark (\url{https://github.com/lark-parser/lark}) for creating the Parser, which is the only required dependence to run it.
Moreover, the tool requires to use the version 3.10 (or higher) of Python.
Concerning the distribution licence, CNL2ASP is released under the Apache 2.0 licence, a permissive open-source licence, which allows the user to use it also in industrial contexts.
Its usage is quite intuitive since it can be used as a standalone tool by issuing the command
\begin{cnl}
    python3 src/main.py input_file [output_file]
\end{cnl}
or, as an alternative, it can be used as a library in other Python projects by simply importing it.

\section{Synthetic use cases}\label{sec:synthetic}
In this section, we present some examples to demonstrate how the language can be used to define well-known combinatorial problems in a natural and easily understandable way. The corresponding translations into ASP are also provided.

\subsection{Graph coloring}\label{sec:graphcoloring}
We begin by presenting an encoding of the graph coloring problem using our CNL.
We recall that the graph coloring problem is the problem of finding an assignment of colors to nodes in a graph such that two adjacent nodes do not share the same color.

\begin{cnl}
$\label{graphcol:cnl:ln:nodes}$A node goes from 1 to 3.
$\label{graphcol:cnl:ln:colors}$A color is one of red, green, blue.
$\label{graphcol:cnl:ln:startedges}$Node 1 is connected to node X, where X is one of 2, 3.
Node 2 is connected to node X, where X is one of 1, 3.
$\label{graphcol:cnl:ln:endedges}$Node 3 is connected to node X, where X is one of 1, 2.
$\label{graphcol:cnl:ln:choice}$Every node can be assigned to exactly 1 color.
$\label{graphcol:cnl:ln:constraint}$It is required that when node X is connected to node Y then node X is not assigned to color C and also node Y is not assigned to color C.
\end{cnl}
One can notice the presence of explicit definition propositions (lines \ref{graphcol:cnl:ln:nodes}--\ref{graphcol:cnl:ln:endedges}), with a ranged definition proposition (line \ref{graphcol:cnl:ln:nodes}) and a list definition proposition (line \ref{graphcol:cnl:ln:colors}), enumerative definition propositions with where clauses (lines \ref{graphcol:cnl:ln:startedges}--\ref{graphcol:cnl:ln:endedges}), a quantified clause (line \ref{graphcol:cnl:ln:choice}) and, lastly, a positive strong constraint (line \ref{graphcol:cnl:ln:constraint}).

The resulting ASP encoding is the following:
\begin{asp}
node(1..3).
color(1,"red"). color(2,"green").   color(3,"blue").
connected_to(1,2).  connected_to(1,3).
connected_to(2,1).  connected_to(2,3).
connected_to(3,1).  connected_to(3,2).
{assigned_to(_X1,_X2): color(_,_X2)} = 1 :-  node(_X1).
:- connected_to(X,Y), assigned_to(X,C), assigned_to(Y,C).
\end{asp}
where each proposition at line $i$ is translated as the rule(s) reported at line $i$ (with $i=1..7$).

\subsection{Hamiltonian path}
The second problem we consider here is the well-known Hamiltonian path problem, which is the problem of finding a path in a graph that visits each node exactly once, starting from a given node.
\begin{cnl}
$\label{ham:cnl:ln:node}$A node goes from 1 to 5.
$\label{ham:cnl:ln:startconn}$Node 1 is connected to node X, where X is one of 2, 3.
Node 2 is connected to node X, where X is one of 1, 4.
Node 3 is connected to node X, where X is one of 1, 4.
Node 4 is connected to node X, where X is one of 3, 5.
$\label{ham:cnl:ln:endconn}$Node 5 is connected to node X, where X is one of 3, 4.
$\label{ham:cnl:ln:choice}$Every node X can have a path to a node connected to node X.
$\label{ham:cnl:ln:aggr1}$It is required that the number of nodes where node X has a path to is equal to 1.
$\label{ham:cnl:ln:aggr2}$It is required that the number of nodes that have a path to node X is equal to 1.
$\label{ham:cnl:ln:recursion}$Node Y is reachable when node X is reachable and also node X has a path to node Y.
$\label{ham:cnl:ln:nodesreachable}$It is required that every node is reachable.
$\label{ham:cnl:ln:constant}$start is a constant equal to 1.
$\label{ham:cnl:ln:startreachable}$Node start is reachable.
\end{cnl}
Line \ref{ham:cnl:ln:node} defines the nodes and lines from \ref{ham:cnl:ln:startconn}--\ref{ham:cnl:ln:endconn} define the connections between nodes.
Then, line \ref{ham:cnl:ln:choice} reports a quantified proposition with an object accompanied by a verb clause, lines \ref{ham:cnl:ln:aggr1} and \ref{ham:cnl:ln:aggr2} report strong constraint propositions with aggregates, line \ref{ham:cnl:ln:recursion} reports a conditional definition clause, line \ref{ham:cnl:ln:nodesreachable} reports a constraint clause with the presence of a quantifier, and line \ref{ham:cnl:ln:constant} defines the constant \lstinline[language=cnl]|start|, which is subsequently used in line \ref{ham:cnl:ln:startreachable}.
The ASP encoding corresponding to the CNL statements is the following:
\begin{asp}
node(1..5).
connected_to(1,2).  connected_to(1,3).
connected_to(2,1).  connected_to(2,4).
connected_to(3,1).  connected_to(3,4).
connected_to(4,3).  connected_to(4,5).
connected_to(5,3).  connected_to(5,4).
{path_to(X,_X1): connected_to(X,_X1)} :- node(X).
:- node(X), #count{_X2: path_to(X,_X2)} != 1.
:- node(X), #count{_X3: path_to(_X3,X)} != 1.
reachable(Y) :- reachable(X), path_to(X,Y).
:- not reachable(_X4), node(_X4).
$\label{ham:asp:ln:reachable}$reachable(1).
\end{asp}
where a CNL statement at line $i$ is represented by the rule(s) at line $i$ with ($i=1..11$), whereas CNL statements reported in lines \ref{ham:cnl:ln:constant} and \ref{ham:cnl:ln:startreachable} are encoded by the rule at line \ref{ham:asp:ln:reachable}.
As an alternative, one could also use the sentence \lstinline[language=cnl]|start is a constant|, and then use the solver options to change the starting node.

\subsection{Maximal clique}
The third problem is the maximal clique problem, which is the problem of finding a clique, i.e., a subset of the nodes of a given graph where all nodes in the clique are adjacent to each other, and the cardinality of the clique is maximal.
\begin{cnl}
$\label{max:cnl:ln:init}$A node goes from 1 to 5.
Node 1 is connected to node X, where X is one of 2, 3.
Node 2 is connected to node X, where X is one of 1, 3, 4, 5.
Node 3 is connected to node X, where X is one of 1, 2, 4, 5.
Node 4 is connected to node X, where X is one of 2, 3, 5.
$\label{max:cnl:ln:end}$Node 5 is connected to node X, where X is one of 2, 3, 4.
$\label{max:cnl:ln:choice}$Every node can be chosen. 
$\label{max:cnl:ln:constraint}$It is required that when node X is not connected to node Y then node X is not chosen and also node Y is not chosen, where X is different from Y.
$\label{max:cnl:ln:weak}$It is preferred with high priority that the number of nodes that are chosen is maximized.
\end{cnl}
where statements from line \ref{max:cnl:ln:init} to line \ref{max:cnl:ln:end} define the input graph.
Then, line \ref{max:cnl:ln:choice} reports a quantified proposition with no object, line \ref{max:cnl:ln:constraint} contains a strong constraint proposition with a comparison on the variables used inside it, and line \ref{max:cnl:ln:weak} reports a weak constraint expressing a maximization preference on the highest priority level.
The resulting ASP encoding is reported in the following:
\begin{asp}
node(1..5).
connected_to(1,2).  connected_to(1,3).
connected_to(2,1).  connected_to(2,3).  connected_to(2,4).  connected_to(2,5).
connected_to(3,1).  connected_to(3,2).  connected_to(3,4).  connected_to(3,5).
connected_to(4,2).  connected_to(4,3).  connected_to(4,5).
connected_to(5,2).  connected_to(5,3).  connected_to(5,4).
{chosen(_X1)} :- node(_X1).
:- not connected_to(X,Y), chosen(X), chosen(Y), X != Y.
:~ #count{_X1: chosen(_X1)} = _X2. [-_X2@1]
\end{asp}
where each CNL proposition at line $i$ is translated as the rule(s) reported at line $i$ (with $i=1..9$).

\section{Real-world use cases}\label{sec:realword}
In this section, we show the usage of the CNL specifications to encode three real-world problems which we previously addressed using plain ASP encodings, namely the Nurse Scheduling Problem (NSP)~(Section~\ref{sec:nsp}; \citep{DBLP:conf/lpnmr/DodaroM17}), the Manipulation of Articulated Objects Using Dual-Arm Robots~(Section~\ref{sec:mao}; \citep{DBLP:journals/tplp/BertolucciCDLMM21}), and the Chemotherapy Treatment Scheduling (CTS) Problem~(Section~\ref{sec:cts}; \citep{DBLP:journals/tplp/DodaroGGMMP21}).
Moreover, for each of the reported problem, we also show an empirical analysis comparing the performance of the encoding generated in an automatic way by CNL2ASP and the encoding written by human experts. The encodings were compared using the same solver, i.e., \textsc{clingo} version 5.6.1 configured with the same options used in the original papers where the problems were presented. The experiments were executed on a AMD Ryzen 5 2600 with 3.4 GHz, with time and memory limits set to 1200 seconds and 8 GB, respectively.
For the sake of the readability, we do not report in this section the generated encodings, which are however available in Appendix A.

\subsection{Nurse scheduling problem (NSP)}
\label{sec:nsp}
The NSP is the problem of computing an assignment of nurses to shifts (morning, afternoon, night, or rest) in a given period of time such that the assignment satisfies a set of requirements.
In particular, the NSP described in this section was originally defined by \citeN{DBLP:conf/lpnmr/DodaroM17}, where authors classified the requirements as follows: \textit{(i)} Hospital requirements, which impose the length of the shifts and that each shift is associated with a minimum and a maximum number of nurses that must be present in the hospital; \textit{(ii)} Nurses requirements, which impose that nurses have a limit on the minimum and maximum number of working hours during the considered period of time, and that each nurse has an adequate rest period; \textit{(iii)} Balance requirements, which impose that the number of times a nurse can be assigned to morning, afternoon and night shifts is fixed. 

The first part of our CNL specifications concerns the definition of the domain and of the input facts of the NSP, and it is reported in the following:
\begin{cnl}
numberOfNurses is a constant.
A nurse goes from 1 to numberOfNurses.
A day goes from 1 to 365 and is made of shifts that are made of hours.
A shift is one of morning, afternoon, night, specrest, rest, vacation and has hours that are equal to respectively 7, 7, 10, 0, 0, 0.
maxNurseMorning is a constant.
maxNurseAfternoon is a constant.
maxNurseNight is a constant.
minNurseMorning is a constant.
minNurseAfternoon is a constant.
minNurseNight is a constant.
maxHours is a constant equal to 1692.
minHours is a constant equal to 1687.
maxDay is a constant equal to 82.
maxNight is a constant equal to 61.
minDay is a constant equal to 74.
minNight is a constant equal to 58.
balanceNurseDay is a constant equal to 78.
balanceNurseAfternoon is a constant equal to 78.
balanceNurseNight is a constant equal to 60.
\end{cnl}
In the definition above, we used implicit definition propositions that therefore also create the input facts of the problem. Note that the number of nurses is a constant that is specified by the user, some constants like \lstinline[language=cnl]|maxNurseMorning|, \lstinline[language=cnl]|maxNurseAfternoon|, etc. depend on the number of nurses, therefore they are also left to the user, whereas all other constants are specific to the NSP considered, therefore they are stated.

Then, the second part of our CNL specifications are used for solving the problem:
\begin{cnl}
Every nurse can work in exactly 1 shift for each day.
It is required that the number of nurses that work in shift S for each day is at most M, where S is one of morning, afternoon, night and M is one of respectively maxNurseMorning, maxNurseAfternoon, maxNurseNight.
It is prohibited that the number of nurses that work in shift S for each day is less than M, where S is one of morning, afternoon, night and M is one of respectively minNurseMorning, minNurseAfternoon, minNurseNight.
It is prohibited that the total of hours in a day where a nurse works in is more than maxHours.
It is prohibited that the total of hours in a day where a nurse works in is less than minHours.
It is prohibited that the number of days with shift vacation where a nurse works in is different from 30.
It is prohibited that a nurse works in shift S in a day and also the next day works in a shift before S, where S is between morning and night.
It is required that the number of occurrences between each 14 days with shift rest where a nurse works in is at least 2.$\label{prop:nsp:required1}$
It is required that when a nurse works in shift night for 2 consecutive days then the next day works in shift specrest.$\label{prop:nsp:required2}$
It is prohibited that a nurse works in a day in shift specrest and also the previous 2 consecutive days does not work in shift night.$\label{prop:nsp:required3}$
It is prohibited that the number of days with shift S where a nurse works in is more than M, where S is one of morning, afternoon, night and M is one of respectively maxDay, maxDay, maxNight.
It is prohibited that the number of days with shift S where a nurse works in is less than M, where S is one of morning, afternoon, night and M is one of respectively minDay, minDay, minNight.
It is preferred, with high priority, that the difference in absolute value between B, and the number of days with shift S where a nurse works in ranging between minDay and maxDay is minimized, where B is one of balanceNurseDay, balanceNurseAfternoon and S is one of morning, afternoon.
It is preferred, with high priority, that the difference in absolute value between balanceNurseNight, and the number of days with shift night where a nurse works in ranging between minNight and maxNight is minimized.
\end{cnl}
Here, it is interesting to observe that the specifications first define that a nurse can work in exactly one shift for each day leaving a free choice about the shift to assign to each nurse, and then they impose some requirements on the assigned shift.
Moreover, note that in general we used negative constraints (i.e., sentences starting with \lstinline[language=cnl]|It is prohibited|), with the exception of the ones at lines \ref{prop:nsp:required1} and \ref{prop:nsp:required2} since we found that this formulation is more natural.
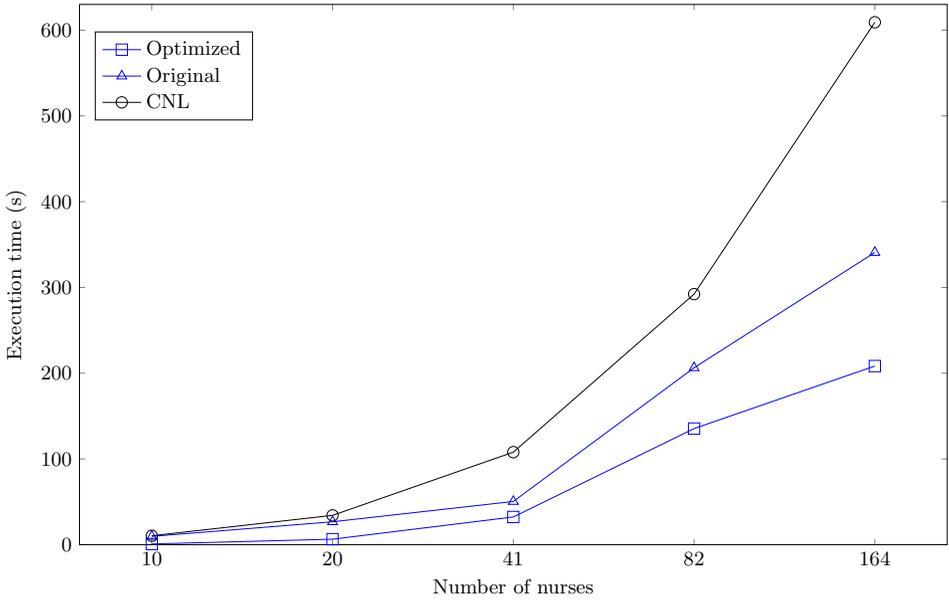
\begin{figure}[t!]
	\centering
		\begin{tikzpicture}[scale=0.85]
		\pgfkeys{%
			/pgf/number format/set thousands separator = {}}
		\begin{axis}[
        	scale only axis
            , legend style = {at={(0.20,0.95)}}
        	, xlabel={Number of nurses}
        	, ylabel={Execution time (s)}
        	, width=1\textwidth
		    , height=0.62\textwidth
        	, ymin=0, ymax=630
        	, symbolic x coords={10,20,41,82,164},
        	, xtick={10,20,41,82,164}
        	, ytick={0,100,200,300,400,500,600}
        	, major tick length=2.5pt        	
        	]				  
         \addlegendentry{Optimized}
		\addplot [mark size=2.5pt, color=blue, mark=square] [unbounded coords=jump] table[col sep=semicolon, y index=3] {./nsp.csv};
        \addlegendentry{Original}
		\addplot [mark size=2.5pt, color=blue, mark=triangle] [unbounded coords=jump] table[col sep=semicolon, y index=1] {./nsp.csv};        
        \addlegendentry{CNL}
		\addplot [mark size=2.5pt, color=black, mark=o] [unbounded coords=jump] table[col sep=semicolon, y index=2] {./nsp.csv};
		\end{axis}
		\end{tikzpicture}
		\caption{Time comparison of the performance of the original, the optimized and the CNL encodings to solve instances of the NSP.\label{fig:nsp}}
\end{figure}
\paragraph{Comparison of the performances.} The encoding generated by the CNL specifications described before has been compared to the original one proposed by \cite{DBLP:conf/lpnmr/DodaroM17} (referred to as \textit{Original}) and with an optimized version proposed by \cite{DBLP:journals/ia/AlvianoDM18} (referred to as \textit{Optimized}).
The experiment consists of five instances of the NSP with increasing number of nurses.
Results are shown in Figure~\ref{fig:nsp}, where it is possible to observe that the Optimized encoding outperforms both the original and the CNL one.
This result is not surprising since the Optimized encoding takes advantage of specific properties of the NSP to reduce the search space for the solver.
Concerning the performance of the CNL encoding, it is possible to observe that it is approximately between 1.5 to 2 times slower than the original one. The main difference in terms of performance is due to the fact that CNL encoding generates aggregates for constraints at lines $\ref{prop:nsp:required2}$ and $\ref{prop:nsp:required3}$, which are less efficient in this context than the normal rules used in the original encoding.
In this respect, tools for the automatic rewriting of aggregates, such as the one proposed by \cite{DBLP:journals/corr/abs-2009-10240}, can be helpful also in our context.
However, it is worth mentioning that, even on the hardest instance, the generated encoding is able to terminate in approximately ten minutes.

\subsection{Manipulation of articulated objects using dual-arm robots}\label{sec:mao}
The manipulation of articulated objects is an important task in real-world robot scenarios.
\citeN{DBLP:journals/tplp/BertolucciCDLMM21} presented an ASP framework for handling the manipulation of an articulated object in a 2D workspace with the possibility of performing actions like rotating one of its link with respect to another one around their joint.
The framework was composed by five modules, namely Knowledge Base, Goal Checker, Consistency Checking, Action Planner, and Motion Planner. All the modules but Motion Planner were implemented using ASP.
In this section, we focus on the Action Planner as described by the encoding reported in the Figure~6 of the paper by \citeN{DBLP:journals/tplp/BertolucciCDLMM21}, since it involves a number of interesting constructs, such as temporal concepts as well as the concept of angle.

The first part of our CNL specifications concerns the definition of the domain of the problem, and it is reported in the following: 
\begin{cnl}
A time is a temporal concept expressed in steps ranging from 1 to 10.
A joint is identified by an id.
An angle is identified by a value.
A position is identified by a joint, by an angle, and by a time.
A link is identified by a first joint, and by a second joint.
A rotation is identified by a first joint, by a second joint, by a desired angle, by a current angle, and by a time.
A goal is identified by a joint, and by an angle.
granularity is a constant.
timemax is a constant.
\end{cnl}
It is possible to observe that we have the concept of time that is marked as temporal. As described in Section~\ref{sec:temporalconcept}, this enables the possibility to use constructs like after, before, and so forth (as shown at line \ref{prop:mao:time} of the second part of the CNL specifications). Moreover, our CNL implicitly handles the concept of angle, e.g., by ensuring that sum operations always create angles whose values are between 0 and 359 degrees.

The second part of the CNL specifications is instead used for solving the problem, and it is reported in the following:
\begin{cnl}
Whenever there is a link with a first joint J1, and with a second joint J2, then we must have a link with a first joint J2, and with a second joint J1.
Whenever there is a time T that is after 0, then we can have at most 1 rotation with a first joint J1, with a second joint J2, with a desired angle A, with a current angle AI, and with time T such that there is a joint J1, a joint J2, an angle A, a link with first joint J1, and with second joint J2, a position with joint J1, with angle AI, and with time T.$\label{prop:mao:time}$
It is required that T is less than timemax, whenever there is a rotation with time T.
It is required that the first joint J1 of the rotation R is greater than the second joint J2 of the rotation R, whenever there is a rotation R with first joint J1, with second joint J2.
It is required that the desired angle A of the rotation R is different from the current angle AI of the rotation R,  whenever there is a rotation R with desired angle A, and with current angle AI.
It is required that the sum between the desired angle A of the rotation R and granularity is equal to the current angle AI of the rotation R, whenever there is a rotation R with desired angle A greater than 0, with current angle AI greater than A.
It is required that the sum between the current angle AI of the rotation R and granularity is equal to the desired angle A of the rotation R, whenever there is a rotation R with current angle AI greater than 0, with desired angle A greater than AI.
It is required that the difference between 360 and granularity is equal to the desired angle A of the rotation R, whenever there is a rotation R with desired angle A, and with current angle equal to 0.
It is required that the difference between 360 and granularity is equal to the current angle AI of the rotation R, whenever there is a rotation R with desired angle A equal to 0, and with current angle AI.
Whenever there is a joint J, whenever there is a time T, then we can have a position with joint J, with angle A, and with time T to exactly 1 angle A.
It is required that the angle A1 of the position P1 is equal to the angle A2 of the position P2, whenever there is a position P1 with joint J, with angle A1, and with time T, whenever there is a position P2 the next step with joint J, and with angle A2, whenever there is not a rotation with time T less than or equal to timemax.
It is required that the angle A1 of the position P is equal to the desired angle A2 of the rotation R, whenever there is a position P with joint J1, with time T, with angle A1, whenever there is a rotation R the previous step with first joint J1, and with desired angle A2.
It is required that the angle AN of the position P is equal to |AC+(A-AP)+360|, whenever there is a time T, whenever there is a position P the next step with joint J1, and with angle AN, whenever there is a rotation with first joint J2, with desired angle A, with current angle AP, and with time T, whenever there is a position P2 with joint J1 greater than J2, with angle AC, and with time T.
It is required that the angle A1 of the position P1 is equal to the angle A2 of the position P2, whenever there is a position P1 with joint J1, with angle A1, and with time T, whenever there is a position P2 with joint J1, and with angle A2, and with the next step respect to T, whenever there is a rotation with first joint J2 greater than J1, and with time T not after timemax.
It is required that the angle A1 of the goal G is equal to the angle A2 of the position P, whenever there is a goal G with joint J, with angle A1, whenever there is a position P with joint J, with angle A2, and with time equal to timemax.
\end{cnl}
Here, due to the structure of the problem, we found more natural to use positive constraints.

\begin{table}[t]
    \centering
    \begin{tabular}{rrrrrrrrr}
    \toprule
    \toprule
&	\multicolumn{2}{c}{12 joints, 180 granularity} && \multicolumn{2}{c}{12 joints, 90 granularity}&&\multicolumn{2}{c}{14 joints, 180 granularity}\\
\cmidrule{2-3}\cmidrule{5-6}\cmidrule{8-9}
Instance &	Original	&CNL	&&	Original	&CNL	&&	Original&	CNL\\
\cmidrule{1-9}
1	&0.01&	0.01	&&	49.7	&8.4	&&	0.03&	0.06\\
2	&0.01&	0.01	&&	0.1	&0.3	&&	0.04&	0.09\\
3	&0.01&	0.01	&&	51.3	&2.8	&&	0.03&	0.03\\
4	&0.01&	0.03	&&	24.8	&4.3	&&	0.01&   0.03\\
5	&0.01&	0.01	&&	20.7	&3.8	&&	0.01&	0.01\\
6	&0.01&	0.01	&&	27.7	&5.8	&&	0.01&	0.04\\
7	&0.01&	0.01	&&	0.1 	&0.3	&&	0.01&	0.01\\
8	&0.01&	0.01	&&	14.7	&1.8	&&	0.01&	0.05\\
9	&0.01&	0.01	&&	0.06	&0.1	&&	0.01&	0.05\\
10	&0.01&	0.01	&&	0.1	&0.2	&&	0.01&	0.03\\
\bottomrule
\bottomrule
    \end{tabular}
    \caption{Comparison of time (in seconds) employed by the original encoding and by the CNL encoding to compute a solution within 10 steps or to prove that there is no solution.}
    \label{tab:mao}
\end{table}
\paragraph{Comparison of the performances.}
The encoding generated by the CNL specifications described before has been compared to the original one proposed by \citeN{DBLP:journals/tplp/BertolucciCDLMM21}, referred to as \textit{Original}.
In particular, we considered all the instances with 12 and 14 joints, with granularity equal to 180 and 90, and we set the number of maximum steps equal to 10.
Such instances represent the biggest ones in terms of number of joints and granularity.
Results are shown in Table~\ref{tab:mao}, where we report, for both the original and the generated encodings, the time (expressed in seconds) for computing a solution within the maximum number of time steps, or to prove that there is no a solution within such a limit.
It is possible to observe that there is no overhead introduced by the CNL encoding, which is actually faster than the original one on some instances.
In particular, we observed that the generated encoding is faster on instances where there is no solution within 10 time steps (i.e., unsatisfiable instances). This difference seems to be related to the structure of the encodings, since the original encoding uses some direct rules to compute the position of joint angles which are not modified in a given time step, whereas the same task is performed by the generated encoding by using a choice rule and some constraints. This structure seems to be heuristically preferred by the solver.

\subsection{Chemotherapy treatment scheduling (CTS) problem}
\label{sec:cts}
The CTS problem is a complex problem taking into account different constraints and resources. In this section, we consider a simplified version of the problem described by \citeN{DBLP:journals/tplp/DodaroGGMMP21} that presented a case study based on the requirements of an Italian hospital.
The idea here is to focus on the main constraints and optimization statements that are useful to show the capabilities of our tool, without considering all the variants described by \citeN{DBLP:journals/tplp/DodaroGGMMP21}.
In particular, the CTS problem consists of assigning a starting hour to the treatment of all the patients, and to the phases before the treatment, where the phases are \textit{(i)} the admission to the hospital, \textit{(ii)} the blood collection, and \textit{(iii)} the medical check.
Moreover, during the treatment, each patient must be assigned either to a bed or a chair.
A proper solution to the CTS problem requires the satisfaction of a number of constraints, e.g.,
the starting time of the admission to the hospital must be after the opening time of the hospital,
patients with long therapy must be assigned after 11:20 AM, 
and each bed or chair must be assigned to just one patient at a time. 
Finally, every patient has a preference between chairs and beds and the solution should try to maximize the number of patients assigned to the preferred resource.

The first part of our CNL specifications concerns the definition of the domain of the problem, and it is reported in the following:
\begin{cnl}
A timeslot is a temporal concept expressed in minutes ranging from 07:30 AM to 01:30 PM with a length of 10 minutes.
A day is a temporal concept expressed in days ranging from 01/01/2022 to 07/01/2022.
A patient is identified by an id, and has a preference.
A registration is identified by a patient, and by an order, and has a number of waiting days, a duration of the first phase, a duration of the second phase, a duration of the third phase, and a duration of the fourth phase.
A seat is identified by an id, and has a type.
An assignment is identified by a registration, by a day, and by a timeslot.
A position in is identified by a patient, by an id, by a timeslot, and by a day.
\end{cnl}

The second part of the CNL defines the CTS problem, and it is reported in the following:
\begin{cnl}
Whenever there is a registration R with an order equal to 0, then R can have an assignment to exactly 1 day, and timeslot.
Whenever there is a registration R with patient P, with order OR, and with a number of waiting days W, whenever there is an assignment with registration patient P, with registration order OR-1, and with day D, whenever there is a day with day D+W, then we can have an assignment with registration R, and with day D+W to exactly 1 timeslot.
It is required that the sum between the duration of the first phase of the registration R, the duration of the second phase of the registration R, and the duration of the third phase of the registration R is greater than the timeslot of the assignment A, whenever there is a registration R, whenever there is an assignment A with registration R, with timeslot T.
Whenever there is a patient P, whenever there is an assignment with registration patient P, with timeslot T, and with day D, whenever there is a registration R with patient P, and with a duration of the fourth phase PH4 greater than 0, then P can have a position with id S, with timeslot T, with day D in exactly 1 seat S for PH4 timeslots.
It is required that the number of patient that have position in id S, day D, timeslot TS is less than 2, whenever there is a day D, whenever there is a timeslot TS, whenever there is a seat with id S.
It is required that the assignment A is after 11:20 AM, whenever there is a registration R with a duration of the fourth phase greater than 50 timeslots, whenever there is an assignment A with registration R.
It is preferred as much as possible, with high priority, that a patient P with preference T has a position in a seat S, whenever there is a seat S with type T.
\end{cnl}
Here, we want to emphasize the simplicity of using specific constructs for temporal concepts like the time slot, as done in the sentence at line 6, where we state that an assignment is after 11:20 AM.

\begin{figure}[t!]
	\centering
		\begin{tikzpicture}[scale=0.85]
		\pgfkeys{%
			/pgf/number format/set thousands separator = {}}
		\begin{axis}[
		scale only axis		
		%, legend style = {draw=none,fill=none,at={(0.45,0.9)}}
		, xlabel={Instance}
		, ylabel={Execution time (s)}
		, width=1\textwidth
		, height=0.62\textwidth
		, ymin=0, ymax=150
		, xmin=0, xmax=22
		, ytick={0,10,20,30,40,50,60,70,80,90,100,110,120,130,140,150}
		, xtick={1,2,3,4,5,6,7,8,9,10,11,12,13,14,15,16,17,18,19,20,21,22} 
		, major tick length=2.5pt
		]				  
        \addlegendentry{Original}
		\addplot [mark size=2.5pt, color=blue, mark=triangle] [unbounded coords=jump] table[col sep=semicolon, y index=1] {./cts.csv};

        \addlegendentry{CNL}
		\addplot [mark size=2.5pt, color=black, mark=o] [unbounded coords=jump] table[col sep=semicolon, y index=2] {./cts.csv};
		\end{axis}
		\end{tikzpicture}
		\caption{Time comparison of the performance of the original and CNL encodings to solve instances of the CTS problem.\label{fig:cts}}
\end{figure}
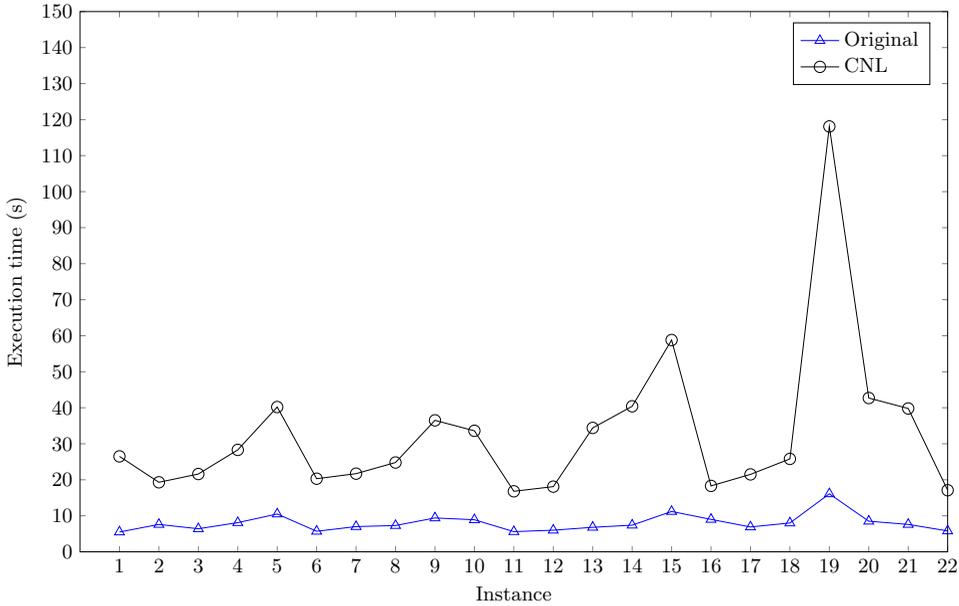
\paragraph{Comparison of the performances.}
The encoding generated by the CNL specifications described before has been compared to the original one proposed by \citeN{DBLP:journals/tplp/DodaroGGMMP21}, referred to as \textit{Original}.
The results are presented in Figure~\ref{fig:cts}. As expected, the original encoding is in general faster than the generated encoding. Nevertheless, the performance of generated encoding is still satisfactory, since on average it requires 32 seconds to compute a solution, with a peak of 2 minutes on the hardest instance.

\section{Preliminary User Validation}\label{sec:uservalidation}
In this section, we present an analysis conducted to assess the usability and readability of the proposed CNL.
The test was conducted on August 1st, 2023, and involved 10 individuals among doctoral students and researchers from the Department of Mathematics and Computer Science at the University of Calabria. It is worth noting that 5 participants work with ASP daily and can be considered experts, while the other 5 work on different research topics. Additionally, 7 participants had attended at least one course on ASP during their studies, whereas the others attended only short seminars about ASP.
The tool was not introduced beforehand, and the content of the experiment was not announced in advance. Moreover, we ensured that:
\textit{(i)} participants had no prior experience with CNL2ASP;
\textit{(ii)} the set of participants did not exclusively consist of individuals interested in tools or those with specific biases toward using programming environments;
\textit{(iii)} the set of participants included a mix of both proficient and less proficient ASP programmers, which is the expected target of users. Indeed, we believe that a limited experience on ASP or at least on declarative languages for solving combinatorial problems might be needed to proficiently use the tool.

Finally, we mention that this analysis should be considered preliminary due to the limited number of participants, and none of them had received prior training on CNL2ASP.

\begin{table}[t!]
    \centering
    \begin{tabular}{clccccc}
    \toprule
    \toprule
        Language	&	Main research area	& Attended ASP course? &	$c_1$	&	$c_2$	&	$c_3$	&	$c_4$\\
        \cmidrule{1-7}
        ASP	&	ASP	Systems and Tools &	Y   &   1	&	1	&	1	&	1\\
        ASP	&	ASP	Systems and Tools &	Y   &   1	&	1	&	1	&	1\\
        ASP	&	ASP Semantics and Theory	&	Y   &   1	&	1	&	1	&	0\\
        ASP	&	Deep Learning &   Y	&	0	&	0	&	0	&	0\\
        ASP	&	Deep Learning &   Y	&	0	&	0	&	0	&	0\\
        \cmidrule{1-7}
        CNL	&	ASP	Systems and Tools &	Y   &   1	&	0	&	1	&	0\\
        CNL	&	ASP	Systems and Tools &	Y   &   0	&	0	&	0	&	0\\
        CNL	&	Theoretical Computer Science  & N	&	0	&	1	&	0	&	0\\
        CNL	&	Deep Learning	&   N   &	0	&	0	&	0	&	0\\
        CNL	&	Deep Learning	&   N   &	0	&	0	&	0	&	0\\        
    \bottomrule
    \bottomrule
    \end{tabular}
    \caption{Results on the usability test. Each row represents the results of an individual participant. A value of '1' indicates that the provided ASP rule/CNL specification was correct, while '0' indicates that it was incorrect.}
    \label{tab:usability}
\end{table}
\subsection{Usability}
We designed a test in which participants were asked to solve the following problem:
Given a set of $n$ persons and $m$ teams (assuming $n > m$), the goal is to assign persons to teams while satisfying the following conditions:
\begin{itemize}
    \item Each person must be assigned to exactly one team ($c_1$).
    \item Each team can have a maximum of 4 persons ($c_2$).
    \item Two persons who are incompatible cannot be on the same team ($c_3$).
    \item If possible, two friends should be placed in the same team ($c_4$).
\end{itemize}

We divided the participants into two groups. The first group was expected to use ASP to solve the problem, while the second group was instructed to use our CNL. The test began with a brief description of the task and some basic instructions on the CNL syntax for the second group. To ensure a fair comparison, individuals who had never attended an ASP course were included in the second group.

The results are presented in Table~\ref{tab:usability}. As expected, participants familiar with ASP were able to create an ASP program that successfully addressed the given problem. In contrast, individuals who had taken an ASP course during their studies but were not actively using ASP were unable to solve the problem.

Regarding the second group, the best performance came from a researcher who also had experience with ASP, achieving partial success in solving the problem. Interestingly, one of the researchers who did not work with ASP managed to correctly specify condition $c_2$. Consequently, even without prior training, 2 out of 5 participants in this group were able to specify some of the problem's conditions accurately.

\begin{table}[t!]
    \centering
    \begin{tabular}{cccccccccccc}
    \toprule
    \toprule
        ASP &	$s_1$	&	$s_2$	&	$s_3$	&	$s_4$	&	$s_5$	&	$s_6$	&	$s_7$	&	$s_8$	&	$s_9$	&	$s_{10}$	&	Number of correct answers\\
        \cmidrule{1-12}        
        ASP & 0	&	1	&	1	&	0	&	1	&	0	&	0	&	0	&	1	&	0	&	4\\
ASP	&	0	&	1	&	1	&	0	&	1	&	0	&	0	&	1	&	1	&	0	&	5\\
ASP	&	0	&	1	&	1	&	1	&	0	&	0	&	0	&	0	&	0	&	0	&	3\\
ASP	&	1	&	1	&	1	&	1	&	1	&	1	&	0	&	1	&	1	&	0	&	8\\
ASP	&	1	&	1	&	1	&	1	&	1	&	1	&	0	&	1	&	1	&	0	&	8\\
\cmidrule{1-12}
CNL &	1	&	1	&	1	&	0	&	1	&	0	&	0	&	1	&	1	&	0	&	6\\
CNL	&	1	&	1	&	1	&	1	&	1	&	1	&	1	&	1	&	1	&	1	&	10\\
CNL	&	1	&	0	&	1	&	1	&	1	&	1	&	1	&	1	&	0	&	1	&	8\\
CNL	&	1	&	1	&	1	&	1	&	1	&	0	&	0	&	1	&	1	&	0	&	7\\
CNL	&	0	&	0	&	0	&	1	&	1	&	1	&	0	&	0	&	0	&	0	&	3\\
    \bottomrule
    \bottomrule
    \end{tabular}
    \caption{Results on the readability test. Each row represents the results of an individual participant. A '1' indicates that the participant correctly identified the truth or falsity of the corresponding statement, while '0' denotes an incorrect or an empty response.}
    \label{tab:readability}
\end{table}
\subsection{Readability}
We designed a test in which participants were required to determine the truth or falsity of the following statements:
\begin{enumerate}
    \item After two consecutive nights there is a special rest day.
    \item Each nurse has at least 2 rest days every two weeks.
    \item Each nurse has exactly 30 days of holidays.
    \item A nurse can work at most two consecutive nights.
    \item Each nurse has at most 30 days of holidays.
    \item A nurse can work at most three consecutive nights.
    \item A special rest day must be provided when a nurse is in vacation.
    \item Each nurse can be assigned to at most ``maxNight" nights shift during the whole year.
    \item Each nurse can be assigned to at least ``minNight" nights shift during the whole year.
    \item Each nurse should be assigned to exactly ``balanceNurseNight" nights shift during the whole year.
\end{enumerate}
Subsequently, we grouped the participants in the same manner as in the usability experiment. The first group was provided with the ASP encoding for the Nurse Scheduling Problem (NSP) as described by \cite{DBLP:conf/lpnmr/DodaroM17}. The second group received the CNL specifications described in Section~\ref{sec:nsp}.
To alleviate social pressure, we requested that participants remain anonymous during this test.
The fifth statement ($s_5$) was contested as ambiguous, as it can be interpreted as both true and false. Therefore, we assigned a score of 1 for both true and false responses and 0 if the answer was left blank.
Results are reported in Table~\ref{tab:readability}.
On average, participants in the CNL group obtained a score of 6.8 with a peak of 10, whereas participants in ASP group obtained a score of 5.6 with a peak of 8.

\section{Related work}
In this section, we overview the main CNLs proposed in the area of logic programming; for a complete review of CNL, we refer the reader to the interesting survey by \cite{DBLP:journals/coling/Kuhn14}.

One of the first attempts of designing encoding expressed in a CNL as logic programs was presented by 
\cite{DBLP:journals/corr/abs-cmp-lg-9507009} and by \cite{DBLP:conf/wlp/SchwitterHF95}, where Attempto CNL~\citep{DBLP:conf/iccs/Fuchs05} was proposed, whose idea was to convert sentences expressed in a CNL as Prolog clauses.
\cite{DBLP:conf/flairs/ClarkHJTW05} presented a Computer-Processable Language (CPL), whose key principle was to be easier for computers rather than a language easier for users.
Moreover, they presented also an interpreter and a reasoner for this language, and discussed the strengths and weaknesses of natural languages to be used as a the basis for knowledge acquisition and representation.

Concerning ASP, \cite{DBLP:conf/bionlp/ErdemY09} proposed BIOQUERYCNL, a CNL for biomedical queries, and developed an algorithm designed to automatically encode a biomedical query expressed in this language as an ASP program.
BIOQUERYCNL is a subset of Attempto CNL and it can represent queries of the form \textit{Which symptoms are alleviated by the drug Epinephrine?} (we refer the reader to Chapter 3 of \citep{DBLP:conf/bionlp/ErdemY09} for more queries).
Later on, BIOQUERYCNL was also used as a basis to generate explanation of complex queries \citep{DBLP:conf/aaai/OztokE11}.
The main difference with our approach is that CNL2ASP does not cover query answering and is not specialized on one particular application context.

\cite{DBLP:conf/kr/BaralD12} proposed a CNL specific for solving logic puzzles.
The CNL was split into two sets of sentences, namely \textit{Puzzle Domain data} and \textit{Puzzle clues}.
The former plays a similar role of our explicit domain definitions (see Section~\ref{sec:explicit}), whereas Puzzle clues can be seen as the logic rules to solve the puzzle.
As in our case, the CNL was then automatically converted into ASP rules.

\cite{DBLP:conf/iclp/Lifschitz22} showed the process of translating the English sentence
``A prime is a natural number greater than 1 that is not a product of two smaller natural numbers."
into executable ASP code.

\cite{DBLP:journals/tplp/Schwitter18} defined the language PENG\textsuperscript{ASP}, a CNL that is automatically converted into ASP.
Albeit some aspects of PENG\textsuperscript{ASP}'s grammar rules are present in the grammar of our CNL, the latter is geared more towards the formal definition of combinatorial problems in a natural-feeling and unambiguous way that is also reliably predictable in its translation to ASP, choosing words that would stand out easily during reading and with an easily deducible meaning from the given context; this meant sacrificing some of the naturalness of PENG\textsuperscript{ASP}.
In addition, the grammar of PENG\textsuperscript{ASP} is designed for allowing a conversion from the CNL to ASP and then back in the other direction, whereas in CNL2ASP this possibility is not yet available, although the language has been designed in such a way that it should be possible to make it viable.
Another feature that is available in PENG\textsuperscript{ASP} is the possibility to express queries, which is not possible in our CNL.
However, our CNL presents some features that, to best of our knowledge, are not available in PENG\textsuperscript{ASP}, such as explicit definitions, and positive strong constraints, that we found to be useful for specifying real-world problems in a natural way.
Moreover, it should be noted that the implementation of PENG\textsuperscript{ASP}, as well as a binary executable, is not yet public, whereas the implementation of CNL2ASP is open source and publicly available.
As an example of the differences with our CNL, we report a comparison with the CNL for specifying the graph coloring problem used by PENG\textsuperscript{ASP} (Figure~5 of \citep{DBLP:journals/tplp/Schwitter18}\footnote{Since PENG\textsuperscript{ASP} is not publicly available we could not compare the two languages on the other problems used in our paper.}).
\begin{cnl}
$\label{cnl:ln:pengasp:first}$The node 1 is connected to the nodes 2 and 3.
The node 2 is connected to the nodes 1 and 3.
The node 3 is connected to the nodes 1 and 2.
Red is a colour. $\qquad$ Green is a colour. $\qquad$ Blue is a colour.
Every node is assigned to exactly one colour.
It is not the case that a node X is assigned to a colour and a node Y is assigned to the colour and the node X is connected to the node Y.
\end{cnl}
There are two main differences with our CNL presented in Section~\ref{sec:graphcoloring}. The first one is that our CNL must use variables (i.e., \lstinline[language=cnl]|X| in our example) also to specify the connections, whereas the one of PENG\textsuperscript{ASP} does not need it. In our case, sentence at line \ref{cnl:ln:pengasp:first} would create the atom \lstinline[language=cnl]{connected_to(1,2,3)}.
%The second difference is that PENG\textsuperscript{ASP} uses \lstinline[language=cnl]|is assigned| to specify a choice rule, whereas our CNL uses \lstinline[language=cnl]|can be assigned|. 
Secondly, the last sentence is expressed in a negative form in case of PENG\textsuperscript{ASP}, which is similar to the concept of constraint in ASP, whereas our CNL uses a positive sentence which is similar to the concept of clause in propositional logic.
Moreover, the PENG\textsuperscript{ASP} and the CNL2ASP methodologies differ in the way the sentences are processed before being unified with the grammar rules. First of all, the grammar rules for PENG\textsuperscript{ASP} are specified with a Definite Clause Grammar (DCG), while in our solution the grammar is defined in Extended Backus-Naur Form (EBNF). Moreover, our tool builds a sort of syntax tree for handling the internal structure of the sentences before rewriting them into ASP. While in PENG\textsuperscript{ASP}, after the collection in the DCG, a chart parser is used to extract the information needed for the translation and this information can be parsed and passed to the users to help with completing the sentence.

\begin{table}[t]
    \centering
    \begin{tabular}{ccccc}
    \toprule
    \toprule
    Characteristic &	CNL2ASP 	& $\lambda$-Based	&	BIOQUERYCNL	& PENG\textsuperscript{ASP} \\
\cmidrule{1-5}
Simple sentences	            & Y & Y & Y & Y\\
Modifying clauses	            & Y & Y & Y & Y\\
Comparative clauses	            & Y & Y & Y & Y\\
Conjunction/disjunction clauses & Y & N & Y & Y\\
Conditional sentences	        & Y & N & N & Y\\
Negated sentences	            & Y & Y & N & Y\\
Cardinality constraints         & Y & N & Y & Y\\
Aggregates	                    & Y & N & N & U\\
Temporal sentences	            & Y & Y & N & Y\\
Preferences	                    & Y & N & N & Y\\
Queries	                        & N & Y & Y & Y\\
\bottomrule
\bottomrule
    \end{tabular}
    \caption{Comparison of the linguistic features of CNL2ASP, $\lambda$-based (\cite{DBLP:conf/kr/BaralD12}), BIOQUERYCNL (\cite{DBLP:conf/bionlp/ErdemY09}), and PENG\textsuperscript{ASP} (\cite{DBLP:journals/tplp/Schwitter18}). Yes (Y) means the construct is supported, No (N) means that the construct is not supported, Unknown (U) means that there is no evidence that the construct is supported nor unsupported.}
    \label{tab:cnls}
\end{table}

We also mention that some of the sentences used in the CNL presented in this paper are inspired by the Semantics of Business Vocabulary and Business Rules (SBVR)~\citep{DBLP:conf/aaaiss/BajwaLB11,sbvr}, which is a standard proposed by the Object Management Group to formally describe complex entities, e.g., the ones related to a business, using natural language.

%% AGGIUNTA 15/09 - Espressività
In Table~\ref{tab:cnls} we present a comparison of the features of the different CNLs translating to ASP. In particular, we want to highlight the constructs that are covered by the CNLs in order to ease the usage of the tool and to be more adherent to natural language. We considered the same constructs considered in~\cite{DBLP:conf/wlp/SchwitterHF95} plus temporal sentences and new constructs specifically adopted for ASP: cardinality constraints, aggregates and preferences. 
%% AGGIUNTA 15/09 - Espressività

\paragraph{Comparison to previous work.} This paper represents an extended version of the paper \citep{DBLP:conf/datalog/DodaroMR22} presented at the 4th International Workshop on the Resurgence of Datalog in Academia and Industry (DATALOG 2.0 2022).
In this paper, the following additional contributions are provided:
\begin{enumerate}
    \item We extended the CNL to support some of the ASP constructs missing in the previous paper, such as disjunctive rules and \lstinline[language=asp]|#min| and \lstinline[language=asp]|#max| aggregates.
    \item We extended the CNL to include explicit definitions and temporal constructs, which we found useful in practice to have a natural description of complex real-world problems.
    \item We extended the CNL with novel constructs such as \lstinline[language=cnl]|whenever/then| clauses that should also make more intuitive the CNL for ASP users.
    \item We included as use cases three additional real-world problems and, for each of them, we executed an experimental analysis comparing the performance of the generated encoding with the one produced by ASP practitioners.
    \item We performed a preliminary user validation, conducted to assess the usability and the readability of the CNL.
    \item We extended the presentation by providing more details about the architecture of CNL2ASP and also by discussing related work.
\end{enumerate}

\section{Conclusion}
In this paper, we proposed a tool for automatically converting English sentences expressed using a controlled language into ASP rules.
Moreover, we provided several examples of combinatorial problems that can be specified using our CNL, and their translations as ASP rules.
Concerning future work, the CNL supported by CNL2ASP might be extended to cover additional constructs and language extensions, e.g., temporal operators~\citep{DBLP:journals/tplp/CabalarDSS20} or Constraint ASP (CASP)~\citep{DBLP:conf/lpnmr/Balduccini11,DBLP:journals/tplp/BanbaraKOS17}, and the tool can be extended to implement a bidirectional conversion, where ASP rules are translated into CNL statements.
Another interesting future work can be the integration of novel constructs, similar to temporal ones, and use them for enabling additional features, such as input validation \citep{DBLP:journals/corr/abs-2202-09626}.
It also worth mentioning that the ASP encoding produced by CNL2ASP can be subject of optimization by using tools like the ones presented by \cite{DBLP:journals/corr/abs-2009-10240}, and by \cite{DBLP:conf/lpnmr/LiuTL22}, which automatically rewrite some of the constructs to improve the performance.
Finally, we recall that the tool presented in this paper as well as all examples and encodings used in this paper are available at \url{https://github.com/dodaro/cnl2asp}.

\noindent
\textbf{Competing interests:} The authors declare none.

\section*{Acknowledgments}
The authors are grateful to the anonymous reviewers for their suggestions.
This work was partially supported 
%by the Italian Ministry of Industrial Development (MISE) under project MAP4ID ``Multipurpose Analytics Platform 4 Industrial Data'', N. F/190138/01-03/X44; 
by Italian Ministry of Research (MUR) 
    under PRIN project PINPOINT ``exPlaInable kNowledge-aware PrOcess INTelligence'', CUP  H23C22000280006, 
    under PNRR project FAIR ``Future AI Research'', CUP H23C22000860006,
    under PNRR project Tech4You ``Technologies for climate change adaptation and quality of life improvement'', CUP H23C22000370006; and by GNCS-INdAM. 

\bibliographystyle{tlplike}
\bibliography{bibliography}
\newpage
\appendix
\section{Encoding generated by our CNL tool for reald-world use cases}
\subsection{Nurse Scheduling Problem}
CNL specifications reported in Section~\ref{sec:nsp} are automatically converted as ASP rules by our tool. The generated encoding is the following:
\begin{asp}
nurse(1..numberOfNurses).
day(1..365).
shift(1,"morning",7).
shift(2,"afternoon",7).
shift(3,"night",10).
shift(4,"specrest",0).
shift(5,"rest",0).
shift(6,"vacation",0).
1 <= {work_in(NURSE,DAY,SHIFT):shift(_,SHIFT,_)} <= 1 :- nurse(NURSE), day(DAY).
:- day(DAY), #count{NURSE: work_in(NURSE,DAY,"morning")} > maxNurseMorning.
:- day(DAY), #count{NURSE: work_in(NURSE,DAY,"afternoon")} > maxNurseAfternoon.
:- day(DAY), #count{NURSE: work_in(NURSE,DAY,"night")} > maxNurseNight.
:- day(DAY), #count{NURSE: work_in(NURSE,DAY,"morning")} < minNurseMorning.
:- day(DAY), #count{NURSE: work_in(NURSE,DAY,"afternoon")} < minNurseAfternoon.
:- day(DAY), #count{NURSE: work_in(NURSE,DAY,"night")} < minNurseNight.
:- nurse(NURSE), #sum{HOURS,DAY: work_in(NURSE,DAY,SHIFT), shift(_,SHIFT,HOURS)} > 1692.
:- nurse(NURSE), #sum{HOURS,DAY: work_in(NURSE,DAY,SHIFT), shift(_,SHIFT,HOURS)} < 1687.
:- nurse(NURSE), #count{DAY: work_in(NURSE,DAY,"vacation")} != 30.
:- shift(SO1,SHIFT1,_), shift(SO2,SHIFT2,_), work_in(NURSE,DAY,SHIFT2), work_in(NURSE,DAY+1,SHIFT1), SO1 < SO2, SO2 >= 1, SO2 <= 3, SO1 >= 1, SO1 <= 6.
:- nurse(NURSE), day(D1), D1 <= 352, #count{D2: work_in(NURSE,D2,"rest"), D2 >= D1, D2 <= D1+13} < 2.
:- day(D1), #count{D2: work_in(NURSE,D2,"night"), D2 >= D1, D2 <= D1+1} = 2, not work_in(NURSE,D1+2,"specrest"), nurse(NURSE).
:- work_in(NURSE,DAY,"specrest"), day(DAY), #count{SHIFT: work_in(NURSE,SHIFT,"night"), SHIFT >= DAY-2, SHIFT <= DAY-1} != 2.
:- nurse(NURSE), #count{DAY: work_in(NURSE,DAY,"morning")} > 82.
:- nurse(NURSE), #count{DAY: work_in(NURSE,DAY,"afternoon")} > 82.
:- nurse(NURSE), #count{DAY: work_in(NURSE,DAY,"night")} > 61.
:- nurse(NURSE), #count{DAY: work_in(NURSE,DAY,"morning")} < 74.
:- nurse(NURSE), #count{DAY: work_in(NURSE,DAY,"afternoon")} < 74.
:- nurse(NURSE), #count{DAY: work_in(NURSE,DAY,"night")} < 58.
:~ nurse(NURSE), #count{DAY: work_in(NURSE,DAY,"morning")} = TOTDAYS, TOTDAYS >= 74, TOTDAYS <= 82, RES = |78 - TOTDAYS|. [RES@3, NURSE]
:~ nurse(NURSE), #count{DAY: work_in(NURSE,DAY,"afternoon")} = TOTDAYS, TOTDAYS >= 74, TOTDAYS <= 82, RES = |78 - TOTDAYS|. [RES@3, NURSE]
:~ nurse(NURSE), #count{DAY: work_in(NURSE,DAY,"night")} = TOTDAYS, TOTDAYS >= 58, TOTDAYS <= 61, RES = |60 - TOTDAYS|. [RES@3, NURSE]
\end{asp}

\subsection{Manipulation of Articulated Objects Using Dual-Arm Robots}
CNL specifications reported in Section~\ref{sec:mao} are automatically converted as ASP rules by our tool. The generated encoding is the following:
\begin{asp}
time(1).  time(2).  time(3).  time(4).  time(5).  time(6).  time(7).  time(8).  time(9).  time(10).  
link(J2,J1) :- link(J1,J2).
0 <= {rotation(J1,J2,A,AI,time(T)):joint(J1), joint(J2), angle(A), link(J1,J2), position(joint(J1),angle(AI),time(T))} <= 1 :- time(T), T > 0.
:- rotation(_,_,_,_,time(T)), T >= timemax.
:- rotation(J1,J2,_,_,time(_)), J1 <= J2.
:- rotation(_,_,A,AI,time(_)), (A)\360 == (AI)\360.
:- rotation(_,_,A,AI,time(_)), (A)\360 > 0, (AI)\360 > (A)\360, (AI)\360 != (A + granularity)\360.
:- rotation(_,_,A,AI,time(_)), (AI)\360 > 0, (A)\360 > (AI)\360, (A)\360 != (AI + granularity)\360.
:- rotation(_,_,A,0,time(_)), (A)\360 != 360 - granularity.
:- rotation(_,_,0,AI,time(_)), (AI)\360 != 360 - granularity.
1 <= {position(joint(J),angle(A),time(T)):angle(A)} <= 1 :- joint(J), time(T).
:- position(joint(J),angle(A1),time(T)), position(joint(J),angle(A2),time(T+1)), not rotation(_,_,_,_,time(T)), T <= timemax, (A1)\360 != (A2)\360.
:- position(joint(J1),angle(A1),time(T)), rotation(J1,_,A2,_,time(T-1)), (A1)\360 != (A2)\360.
:- time(T), position(joint(J1),angle(AN),time(T+1)), rotation(J2,_,A,AP,time(T)), position(joint(J1),angle(AC),time(T)), J1 > J2, (AN)\360 != (|AC+(A-AP)+360|)\360.
:- position(joint(J1),angle(A1),time(T)), position(joint(J1),angle(A2),time(T+1)), rotation(J2,_,_,_,time(T)), J2 > J1, T <= timemax, (A1)\360 != (A2)\360.
:- goal(joint(J),angle(A1)), position(joint(J),angle(A2),time(timemax)), (A1)\360 != (A2)\360.

\end{asp}
We want to emphasize here that each angle is automatically followed by \verb|\ 360| which represents the module operation to ensure that an angle is always between 0 and 359 degrees.

\subsection{Chemotherapy Treatment Scheduling Problem}
CNL specifications reported in Section~\ref{sec:cts} are automatically converted as ASP rules by our tool. The generated encoding is the following:
\begin{asp}
timeslot(1,"07:30"). timeslot(2,"07:40"). timeslot(3,"07:50").
timeslot(4,"08:00"). timeslot(5,"08:10"). timeslot(6,"08:20"). 
timeslot(7,"08:30"). timeslot(8,"08:40"). timeslot(9,"08:50").  
timeslot(10,"09:00"). timeslot(11,"09:10"). timeslot(12,"09:20").  
timeslot(13,"09:30"). timeslot(14,"09:40"). timeslot(15,"09:50").  
timeslot(16,"10:00"). timeslot(17,"10:10"). timeslot(18,"10:20").  
timeslot(19,"10:30"). timeslot(20,"10:40"). timeslot(21,"10:50").
timeslot(22,"11:00"). timeslot(23,"11:10"). timeslot(24,"11:20").
timeslot(25,"11:30"). timeslot(26,"11:40"). timeslot(27,"11:50").  
timeslot(28,"12:00"). timeslot(29,"12:10"). timeslot(30,"12:20"). 
timeslot(31,"12:30"). timeslot(32,"12:40"). timeslot(33,"12:50"). 
timeslot(34,"13:00"). timeslot(35,"13:10"). timeslot(36,"13:20").
day(1,"01/01/2022").
day(2,"02/01/2022"). 
day(3,"03/01/2022").
day(4,"04/01/2022").
day(5,"05/01/2022").
day(6,"06/01/2022").
day(7,"07/01/2022").
1 <= {assignment(registration(patient(R),0), day(D), timeslot(TS)) : day(D,_), timeslot(TS,_) } <= 1 :- registration(patient(R),0,_,_,_,_,_).
1 <= {assignment(registration(patient(P),OR),day(D+W),timeslot(T)) : timeslot(T,_)} <= 1 :- registration(patient(P),OR,W,_,_,_,_), assignment(registration(patient(P),OR-1), day(D), timeslot(_)), day(D+W,_).
1 <= {position_in(patient(P),S) : seat(S,_)} <= 1 :- patient(P,_), assignment(registration(patient(P),_), day(D), timeslot(T)), registration(patient(P),_,_,_,_,_,PH4), PH4 > 0.
position_in(patient(P),S,timeslot(T), day(D)) :- _generated_support(patient(P),S,timeslot(T),day(D)), position_in(patient(P),S).
_generated_support(patient(P),S,timeslot(T..T+PH4-1),day(D)) :- position_in(patient(P),S), patient(P,_), assignment(registration(patient(P),_), day(D), timeslot(T)), registration(patient(P),_,_,_,_,_,PH4).
:- registration(patient(P),OR,_,PH1,PH2,PH3,_), assignment(registration(patient(P),OR),day(_),timeslot(T)), T <= PH1 + PH2 + PH3.
:- day(D,_), timeslot(TS,_), seat(S,_), #count{P: position_in(patient(P),S,timeslot(TS),day(D))} >= 2.
:- registration(patient(P),OR,_,_,_,_,PH4), assignment(registration(patient(P),OR),day(_),timeslot(TS)), PH4 > 50, TS < 24.
:~ seat(S,T), patient(P,T), position_in(patient(P),S,timeslot(_),day(_)). [-1@3, P]
\end{asp}

\end{document}